\documentclass[%
 aip,
cp,  % Conference Proceedings
 amsmath,amssymb,%nobibnotes,
 reprint,%
]{revtex4-2}

\usepackage{graphicx}% Include figure files
\usepackage{dcolumn}% Align table columns on decimal point
\usepackage{bm}% bold math
\usepackage[main=english,ngerman]{babel}

\usepackage[utf8]{inputenc}
\usepackage[T1]{fontenc}
\usepackage{mathptmx} 

\begin{document}

\title{Generalization of Self-Supervised Vision Transformers for Protein Localization Across Microscopy Domains}% Force line breaks with \\

\author{Ben Isselmann} % Write as First name Surname
 \email{ben.isselmann@h-da.de}
\author{Dilara Göksu}
 \email{dilara.goeksu@stud.h-da.de}
\affiliation{%
Hochschule Darmstadt, Schoefferstraße 3, 64295 Darmstadt,
Germany
%Second institution and/or address% Force line breaks with \\ if necessary
}%
\author{Andreas Weinmann}
 \email{andreas.weinmann@thws.de}

\affiliation{Technische Hochschule Würzburg-Schweinfurt, Ignaz-Schön Strasse 11, 97421 Schweinfurt, Germany}

\date{\today} % It is always \today, today, but any date may be explicitly specified
              % Not printed for conference proceedings

\begin{abstract}
Task-specific microscopy datasets are often small, making it difficult to train deep learning models that extract robust and meaningful features. While recent advances in self-supervised learning (SSL) have shown promise, particularly through pretraining on large, domain-specific datasets, the generalizability of these models across datasets with different staining protocols, imaging modalities, and channel configurations remains largely unexplored. Yet, this cross-domain transferability is essential, especially for small-scale studies that cannot afford to generate large annotated datasets.

Building on this background, we investigated whether DINO-pretrained Vision Transformers, trained on either natural images or domain-specific microscopy data, can generalize effectively to a downstream protein localization task on a distinct dataset.

To explore the generalizability of pretrained DINO weights on task-specific microscopy datasets, we generated embeddings using three DINO models pretrained on ImageNet-1k, HPA, and OpenCell data, respectively. We then assessed the quality of these embeddings by supervised training of a classification head to predict protein localization on the OpenCell dataset.

Our results demonstrate the transferability of self-supervised DINO models pretrained on large datasets such as HPA and ImageNet-1k. Notably, the model pretrained on the microscopy-specific HPA dataset achieved the highest test performance, with a mean macro~$F_1$ score of 0.8221~($\pm$0.0062). This slightly improves even upon the task-specific DINO model trained directly on OpenCell data, which reached 0.8057~($\pm$0.0090). These results underscore the effectiveness of large-scale pretraining and suggest that domain-specific models can generalize well to related but distinct microscopy datasets.

These findings suggest that self-supervised methods like DINO, when pretrained on large, domain-relevant datasets, can enable effective use of deep learning–derived features for fine-tuning on small, task-specific microscopy datasets.

\end{abstract}

\maketitle

\section{Introduction}

The analysis of human cells provides a crucial window into the molecular mechanisms underlying diseases and enables the identification of potential drug targets. Advances in high-throughput technologies have made it possible to investigate cellular function at an unprecedented scale \cite{jia_high-throughput_2022, navin_future_2011, buenrostro_single-cell_2015, bandura_mass_2009, wang_single-cell_2020}. In particular, transcriptomic profiling offers critical insights into gene function and regulation by analyzing RNA expression patterns, complementing genomics, which focuses on identifying mutations and structural variations \cite{heumos_best_2023}. However, these molecular approaches lack spatial resolution and morphological context. This limitation has led to the rise of image-based profiling, which enables the quantification of complex phenotypes and the assessment of treatment effects at the cellular level \cite{mattiazzi_usaj_high-content_2016, grys_machine_2016, scheeder_machine_2018}. Advances in high-throughput microscopy and multiplex staining protocols now allow systematic study of subcellular structures and protein localization across large-scale perturbation screens \cite{reicher_pooled_2024, vincent_phenotypic_2022, boyd_harnessing_2020}.

A central challenge in image-based profiling is the generation of meaningful and robust feature representations. While traditional bio-imaging tools provide handcrafted features, their interpretability comes at the cost of flexibility and scalability. Deep learning (DL) offers a powerful alternative by learning task-relevant representations directly from raw image data, reducing the need for manual feature engineering \cite{lecun_deep_2015, premkumar_single-cell_2024}. Yet, DL models often rely on large annotated datasets, which are costly and limited in scope. This constraint has motivated the exploration of self-supervised learning (SSL), where models learn representations without human labels \cite{kim_self-supervision_2025, pham_cha-maevit_2025, kraus_masked_2024, doron_unbiased_2023}. SSL approaches have demonstrated promise in bio-imaging, but their ability to generalize across different staining protocols and imaging modalities remains underexplored.

This study contributes to addressing this gap by evaluating whether transformer-based SSL backbones (DINO) pretrained on either ImageNet-1k or domain-specific microscopy data can generalize to a downstream protein localization task in the OpenCell dataset. More precisely:

\begin{itemize}
\item We systematically investigate two embedding strategies, channel replication and channel mapping, and demonstrate their impact on transfer performance across DINO backbones pretrained on ImageNet and HPA data.
\item We generate feature embeddings for OpenCell data using each (pre)trained DINO model and evaluate them on a downstream protein localization task.
\item We show that DINO models pretrained on domain-specific data (HPA) and general-purpose data (ImageNet) perform competitively with task-specific training on the OpenCell dataset.

\end{itemize}

\section{Related Work}
Driven by the motivation to leverage microscopy-based approaches, researchers developed the Cell Painting protocol \cite{gustafsdottir_multiplex_2013, bray_cell_2016}, a high-throughput staining method designed to extract rich, quantitative cell profiles. This assay employs a multiplex staining approach, using six fluorescent dyes to label eight cellular components or organelles. The stained and fixed cells are imaged across five microscopy channels. The recently optimized protocol \cite{cimini_optimizing_2023} highlights cell painting as the state-of-the-art assay for morphological profiling and provides a robust framework for the systematic analysis of cellular phenotypes. 

Since its initial introduction, the Cell Painting assay has inspired both academia and the pharmaceutical industry to generate several publicly available datasets, such as the Broad Bioimage Benchmark Collection (BBBC) \cite{ljosa_correction_2013, bray_dataset_2017, caicedo_cell_2022}. Most notably, the JUMP-CP consortium recently released the largest publicly accessible Cell Painting dataset to date, comprising images from over 116,000 chemical perturbations \cite{chandrasekaran_jump_2023}. While Cell Painting focuses on broad morphological profiling through organelle-level staining, other efforts have extended this idea toward protein-specific localization. Among them, the Human Protein Atlas (HPA) \cite{thul_subcellular_2017} adopts a more targeted approach using antibody-based immunofluorescence. Each HPA image includes three fixed reference channels: DAPI for the nucleus, $\beta$-tubulin as a cytoskeletal marker for microtubules, and calreticulin to label the endoplasmic reticulum (ER). These markers provide structural context for a fourth channel, which visualizes the protein of interest (POI) through antibody-based staining. In addition to the Human Protein Atlas (HPA), OpenCell \cite{cho_opencell_2022} is a large-scale imaging project with the goal of providing a high-confidence, artifact-free map of protein localization in human cells. Unlike antibody-based approaches, OpenCell uses CRISPR-mediated genome editing to endogenously tag proteins with fluorescent labels, enabling direct visualization of native protein distribution within cells. Alongside the protein of interest (POI), the nucleus is visualized using Hoechst 33342 staining in a second imaging channel.

The utility of these datasets ultimately depends on how well informative features can be extracted from the images. To this end, the research community has developed several feature engineering approaches, many of which are implemented in bio-imaging software such as CellProfiler \cite{carpenter_cellprofiler_2006}. Complementing these tools, a comprehensive workflow outlined by Caicedo et al. \cite{caicedo_applications_2016} addresses common challenges encountered during the extraction of cell-based features. This workflow provides systematic solutions for handling variability in imaging data, enabling more reliable and reproducible analysis.

The robustness and reproducibility of the entire workflow depend heavily on two critical steps. First, the segmentation of subcellular compartments plays a pivotal role in downstream analysis, as the quality of extracted features is highly dependent on the accuracy of this step. Any segmentation errors can significantly impact the final results. Second, the mitigation of technical site effects such as batch effects, plate layout inconsistencies, and other environmental factors remains essential for ensuring the reliability of the analysis across different experimental
conditions.
The integration of Deep Learning (DL) into microscopy-based profiling emerges in recent studies \cite{pratapa_image-based_2021, tang_morphological_2024}. General, DL leverages neural networks to autonomously learn feature representations directly from raw data, minimizing or even eliminating the need for manual feature engineering \cite{lecun_deep_2015}. In single-cell analysis (SCA), DL enhances the prediction of cell behavior and function, uncovering hidden patterns that aid in disease detection \cite{premkumar_single-cell_2024}. Deep learning models such as U-Nets \cite{ronneberger_u-net_2015}, transformer-based architectures, and foundation models like the Segment Anything Model (SAM) \cite{horst_cellvit_2023} have significantly advanced cell segmentation \cite{minaee_image_2022, alahmari_review_2024, wang_systematic_2024}. 
Among these, Stringer et al. introduced Cellpose \cite{stringer_cellpose_2020}, a deep learning-based framework specifically designed for segmenting cell bodies, membranes, and nuclei. Subsequent versions of Cellpose have incorporated a human-in-the-loop pipeline for rapid prototyping of custom models \cite{pachitariu_cellpose_2022}. Cellpose was extended by including models for image restoration tasks such as de-blurring, denoising, and rescaling, thereby improving segmentation performance and reliability \cite{stringer_cellpose3_2024}. Recently, Cellpose-SAM \cite{cellpose_sam_2024} has been introduced, combining Cellpose with the Segment Anything Model (SAM) \cite{kirillov2023segment} to enhance segmentation accuracy, particularly for challenging or ambiguous cellular structures.

While segmentation plays a critical role in preprocessing and spatial analysis, downstream tasks such as protein localization prediction pose a distinct set of challenges. These tasks often require capturing subtle patterns in fluorescence intensity and spatial organization across multiple imaging channels. One of the most well-known efforts in this domain is the Kaggle Human Protein Atlas classification challenge \cite{ouyang_analysis_2019, le_analysis_2022} which focused on predicting protein localization classes from multi-channel immunofluorescence images. Top-performing teams relied on deep supervised learning models trained on annotated data, using pipelines that combined convolutional neural networks (CNNs), transformer-based encoders, and specialized post-processing strategies. These approaches showcased the potential of task-specific architectures to learn protein-specific features directly from raw image data.

While these supervised approaches demonstrated impressive performance on the Kaggle challenge, they typically rely on large amounts of annotated data, carefully curated pipelines, and are often limited in their ability to generalize beyond the dataset they were trained on. These constraints have motivated the exploration of self-supervised learning (SSL) methods, which aim to learn informative representations without the need for manual labels. In this context, recent studies have evaluated whether SSL models, particularly those based on transformer architectures, can capture biologically meaningful patterns in microscopy images. Among them, Doron et al. \cite{doron_unbiased_2023} evaluated a self-supervised method known as DINO (self-distillation with no labels) \cite{caron_emerging_2021}, employing a Vision Transformer (ViT)~\cite{dosovitskiy_image_2021} as an encoder to extract biologically meaningful features across three publicly available imaging datasets with diverse biological focuses. Among other findings, they observed that the resulting feature space for subcellular protein localization closely resembled that derived from hand-crafted features. While they successfully extracted biologically significant features, the attenuation of confounding factors remains an open challenge. To address this issue, Yao et al. \cite{yao_weakly_2024} proposed combining weakly supervised learning techniques, using perturbations as weak labels, with a cross-batch sampling
strategy. Their approach involved using a set of single cell images with perturbation $p$ for both the teacher and student encoders, but sourced from different batches b and $b'$, thus mitigating batch-specific biases and improving feature generalizability. To date, it remains unclear whether such models can generalize across image sources with differing staining strategies and varying channel compositions, such as the HPA and OpenCell datasets. Addressing this, in this work, we investigate whether DINO-pretrained Vision Transformers can effectively transfer learned feature representations to a downstream task across biologically and technically distinct domains.

\section{Materials and Methods}

\subsection{Datasets}
We used three DINO backbones to evaluate feature extraction on microscopic data. One model was trained by us on the OpenCell dataset, while the other two models were based on publicly available pretrained weights, one trained on ImageNet-1k and the other on a subset of the HPA dataset used in the HPA Kaggle competition. To assess their performance, we used the OpenCell dataset as input to generate embeddings with each backbone, which were then used to predict protein localization. An overview of the datasets used is provided in Table~\ref{tab:comparison_datasets}.

To represent a large and diverse set of natural images, we used a DINO backbone pretrained on ImageNet-1k \cite{russakovsky2015imagenetlargescalevisual}. ImageNet-1k is a well-established benchmark in computer vision, comprising over 1.2 million training images, 50,000 validation images, and 100,000 test images spanning 1,000 object categories. Originally developed for the ImageNet Large Scale Visual Recognition Challenge (ILSVRC), it is widely used for pretraining deep learning models for image classification and object detection. 

In the OpenCell \cite{cho_opencell_2022} dataset, the authors investigated the localization and interactions of human proteins to map the architecture of the human proteome. They generated a library of HEK293T cell lines by using CRISPR to insert fluorescent tags (split-mNeonGreen2) into 1,310 individual proteins. A total of 6,301 two-channel images (tagged protein and nucleus) were acquired. Protein localization was manually annotated across 17 subcellular compartments using a three-tier grading system: Grade 3 indicates prominent localization, Grade 2 denotes less pronounced localization, and Grade 1 represents weak localization patterns.

\citeauthor{doron_unbiased_2023} utilized a subset of the Human Protein Atlas (HPA) Cell Atlas \cite{ouyang_analysis_2019}, which was originally part of the Kaggle competition titled "Human Protein Atlas Image Classification." The competition challenged participants to predict protein localization patterns based solely on image data. The dataset comprises 117,882 images from 35 distinct cell lines, with a roughly balanced distribution across cell types. Each image contains four channels: the antibody-stained target protein, microtubules, the nucleus, and the endoplasmic reticulum. The proteins of interest were categorized into 28 expert-annotated localization classes. Notably, each image was assigned between 1 and 6 labels, and the dataset exhibits significant class imbalance across the 28 protein localization categories.

\begin{table}
\caption{\label{tab:comparison_datasets}ImageNet-1k, OpenCell, and HPA Image Classification datasets.}
\begin{ruledtabular}
\begin{tabular}{llll}
& ImageNet-1k \cite{russakovsky2015imagenetlargescalevisual} & OpenCell \cite{cho_opencell_2022} & HPA Image Classification \cite{thul_subcellular_2017}\\
\hline
Images & 1.2 million & 6,301 & 117,882 \\
Channels & 3 (RGB images) & 2 (protein, nucleus) & 4 (protein, microtubules,\\
& & & nucleus, endoplasmic reticulum) \\
Labels & 1000 object categories & 17 localization labels available & 28 expert-annotated  \\
 &  & for 6120 images, (grades 1–3), & localization classes, \\
& & Single-protein localization & Compartment-level \\
& &  per cell (OpenCell ontology) &  annotations per image/cell \\
& & only HEK293T cell line & 35 distinct cell lines \\
Domain & Natural images & Fluorescence microscopy & Fluorescence microscopy \\
Microscope & - & Spinning disk confocal & Leica SP5 \\
 &  & (Andor Dragonfly) & point-scanning confocal \\
Objective & - & 63× / 1.47 NA oil immersion & 63× / 1.40 NA oil immersion \\
Pixel size & -  & 0.102µm (102.4nm) & 0.08µm (80nm) \\
Lateral resolution  & - & $\sim$ 211nm & $\sim$ 222nm\\
Axial resolution  & - & $\sim$ 750–850nm & $\sim$ 785nm \\
Bit depth  & - & 16-bit & 16-bit\\
Environmental control  & - & Live imaging @ 37°C, 5\% CO$_2$ & Fixed cells (immunostaining) \\
Cell state  & - & Live cells & Fixed, immunolabeled cells \\
Typical cells per image  & - & $\sim$ 10–30 cells per field & $\sim$ 10–30 cells per field \\
Well format  & - & 96-well plates & Glass-bottom chamber slides \\
 &  &  & or multiwell plates \\
Image size  & - & Typically 1024×1024 pixels, with & Typically 2048×2048 px\\
&  & cropped fields of view (FOVs) at 600×600 pixels & \\
% URL & \url{https://www.image-net.org/} & \url{https://opencell.czbiohub.org/} & \url{https://www.proteinatlas.org/} \\
\end{tabular}
\end{ruledtabular}
\end{table}

\subsection{Embedding Methods} \label{embeddings}
To apply DINO models pretrained on ImageNet-1k and HPA to OpenCell data, it is essential to account for differences in channel composition across datasets. As shown in Table~\ref{tab:comparison_datasets}, OpenCell images contain two channels, target protein and nucleus, whereas the HPA dataset includes two additional channels representing microtubules and the endoplasmic reticulum. In contrast, ImageNet-1k consists of standard RGB images, which are not directly aligned with fluorescence microscopy data. Since DINO models expect input formats consistent with their pretraining data, we explored two strategies to address this mismatch: channel replication and channel mapping \cite{Pawlowski085118, chen2024chammibenchmarkchanneladaptivemodels}.
\begin{enumerate}
    \item \textbf{Channel replication:} We used the pretrained DINO to extract features for each channel independently. Afterwards, the individual feature vectors were concatenated. The advantage was that no additional training was required. However, the computational cost and the dimensionality of the final feature vectors scaled linearly with the number of channels.
    \item \textbf{Channel mapping:} We mapped corresponding channels between datasets while padding missing channels with zeros. When mapping OpenCell data to DINO pretrained on HPA, we aligned the protein and nucleus channels directly, while representing microtubules and endoplasmic reticulum with blank channels. Similarly, for OpenCell to ImageNet-1k-pretrained DINO, we mapped the protein channel to the red (R) channel and the nucleus channel to the green (G) channel. The advantage of this approach was the potential reuse of channel-specific features. However, the effectiveness depended on the compatibility of the channel semantics between datasets.
\end{enumerate}

\subsection{DINO}
DINO \cite{caron_emerging_2021} is a self-supervised framework, used to train an unbiased feature extractor.
The key idea is to enforce consistency between the outputs of a student and a teacher network, given different augmented views of the same image. This is achieved without the use of labels, through a self-distillation objective.
Let $\theta_s$ be the parameters of a student network $g_{\theta_s}$ and $\theta_t$ the parameters of a teacher network $g_{\theta_t}$. Furthermore, let $x$ be an input image. Then, the student’s output probability distributions $P_s$ over $K$ dimensions is calculated by normalizing the output of the network $g$ with a softmax function:

\begin{eqnarray}
P_s(x)^{(i)} = \frac{\exp(g_{\theta_s}(x)^{(i)}/\tau_s)}{\sum_{k=1}^{K}\exp(g_{\theta_s}(x)^{(k)}/\tau_s)}, \qquad 1 \leq i \leq K.
\end{eqnarray}
Here, $\tau_s$ is a temperature parameter that controls the smoothness of the output distribution. The teacher's output $P_t$ is calculated analogously replacing $\theta_s$ by $\theta_t$ and $\tau_s$ by $\tau_t$.

Both networks share the same Vision Transformer (ViT) \cite {dosovitskiy_image_2021} architecture which serves as the backbone for the DINO framework. The goal is to match the output of the student network $g_{\theta_s}$ to that of the teacher network $g_{\theta_t}$, by minimizing the cross-entropy loss with respect to the parameters of the student network $\theta_s$, i.e., minimizing $  H(P_t(x), P_s(x)) := - P_t(x) \log P_s(x)$, where on the right hand side the $\log$ is applied component-wise followed by the computation of a scalar product.  Specifically, a set of different views $V$ for the input image  $x$ is generated to obtain invariance to different augmentations. $V$ contains two global views $x_1^g, x_2^g$ and several local, smaller views. While the student processes every view in $V$, the teacher only sees the global views. This asymmetric setting prevents collapse and improves stability during training. The student's parameters $\theta_s$ are learned by minimizing 
\begin{eqnarray}
\min_{\theta_s} \sum_{x \in \{x_1^g, x_2^g\}} \; \sum_{\substack{x' \in V \\ x' \neq x}} H(P_t(x), P_s(x'))
\end{eqnarray}
using stochastic gradient descent. Here, the teacher's parameters $\theta_t$ are frozen. To update the weights of the teacher, an exponential moving average (EMA) on the student’s weights, with the update rule 
\begin{eqnarray}
    \theta_t \leftarrow \lambda \theta_t + (1 - \lambda)\theta_s
\end{eqnarray}
was used where $\lambda$ follows a cosine schedule during training, see \cite{caron_emerging_2021} for details.

During training, both networks include a projection head on top of the backbone. This head is typically a multi-layer perceptron (MLP), and it maps the backbone’s output to a space where the self-distillation loss is applied. These heads are discarded after training. At inference time, we used only the backbone of the teacher network as a feature extractor.

\subsection{Training procedure}
To predict protein localization labels for the OpenCell dataset, we adopted the two-stage approach proposed by \citeauthor{doron_unbiased_2023} \cite{doron_unbiased_2023}. In the first stage, we extracted feature embeddings from each OpenCell image using a (pre)trained DINO backbone. In the second stage, we trained a separate classifier head on each set of embeddings to predict the final protein localization labels.

\subsubsection{Feature Extraction with DINO}
We explore three datasets for (pre)training DINO to generate embeddings of OpenCell images:
\begin{itemize}
    \item \textbf{Natural Images (ImageNet-1k):} The DINO backbone pretrained on ImageNet-1k was originally trained in the initial DINO paper \cite{caron_emerging_2021}. We use the corresponding weights.
    
    \item \textbf{Microscopic Dataset (HPA FOV):} The authors in \cite{doron_unbiased_2023} trained various DINO backbones from scratch on a subset of the HPA dataset at both the FOV and single-cell level. In contrast to the original DINO setup for natural images with three RGB channels, they adapted the architecture to process four-channel fluorescence microscopy images. The pretrained weights are publicly available in the corresponding GitHub repository.
    
    \item \textbf{Downstream Task Dataset (OpenCell):} We trained DINO from scratch on the OpenCell dataset. Images were processed through an augmentation pipeline including random resized crops (global: 224px, scale 0.4-1.0; local: 96px, scale 0.05-0.4), cell warping, and protein rescaling (p=0.2), with additional spatial transformations (50\% flip probability). The model is trained with an AdamW optimizer. Additionally, training duration (epochs) are provided in Table \ref{tab:results}. For self-supervised pretraining, we trained on 90\% of the available data, holding out 10\% as a test set for evaluation.
\end{itemize}

\subsubsection{Classifier Training}
Given the (pre)trained DINO backbones based on the three different datasets described in the previous subsection, we constructed a simple classification head to evaluate their performance on the OpenCell dataset. To address class imbalance, we first resampled each data point with a probability inversely proportional to the frequency of its rarest label. Feature vectors were standardized to have zero mean and unit variance, using statistics computed from the training set; the same normalization parameters were then applied to the validation and test sets.

A multi-layer perceptron (MLP) classifier consisting of three layers:
\begin{enumerate}
\item \textbf{Input linear layer:} 512 hidden units, ReLU activation, dropout regularization ($p = 0.5$);
\item \textbf{Hidden linear layer:} 256 hidden units, ReLU activation, dropout regularization ($p = 0.5$);
\item \textbf{Output linear layer:} 17 output units corresponding to the protein localization classes.
\end{enumerate}

Each classifier head was optimized using the AdamW optimizer (weight decay $= 0.04$, $\beta_1 = 0.9$, $\beta_2 = 0.999$) with an initial learning rate of $10^{-4}$. We trained for 300 epochs with a batch size of 512, employing a cosine annealing learning rate scheduler. The loss function used was binary cross-entropy with logits.

To ensure robust evaluation, we conducted five-fold cross-validation on the same 90\% training split used during self-supervised pretraining, while keeping the original 10\% test set untouched. This setup allowed us to validate model performance across different data partitions, reserving the held-out test set for final evaluation.

\subsubsection{Execution Environment}
All experiments are performed on GPU hardware. The execution environment is described in Table \ref{tab:hardware} in detail.
\begin{table}
\caption{\label{tab:hardware}Hardware specifications for the experimental setup.}
\begin{ruledtabular}
\begin{tabular}{ll}
Hardware Component & Specification \\
\hline
CPU & Intel Xeon Gold 6526Y, 32 CPUs (single-core sockets) @ 2.80 GHz \\
GPU & 2 $\times$ NVIDIA L40S, 2 $\times$ NVIDIA H100 NVL \\
OS & Ubuntu 22.04.5 LTS (64-bit, kernel 5.15.0-141-generic) \\
Python version & 3.10 \\
% PyTorch version & 2.2.2+cu121 \\
CUDA version &  11.5 \\
PyTorch version & 2.2.2 \\
\end{tabular}
\end{ruledtabular}
\end{table}

\subsection{Evaluation}
Because we solved a multi-label classification problem, where each data point can be assigned to zero, one, or several non-overlapping classes, we used the macro-averaged $F_1$ \cite{jurafsky2025speech} score as the evaluation metric. First, we restate the notions of precision and recall for each class $i$ which were defined as 

\begin{eqnarray}
\text{precision$_i$} = \frac{\text{tp$_i$}}{\text{tp$_i$} + \text{fp$_{i}$}}, \qquad \text{recall$_{i}$} = \frac{\text{tp$_i$}}{\text{tp$_i$} + \text{fn$_{i}$}}.
\end{eqnarray}

Here, $\text{tp}_i$ represents the number of true positives for class $i$, while $\text{fn}_i$ and $\text{fp}_i$ denote the numbers of false negatives and false positives, respectively.
Next, the $F_1$ score for class $i$ is defined as the harmonic mean of precision and recall, i.e., 
\begin{eqnarray}
F_{1i} = 2 \cdot \frac{\text{precision$_i$} \cdot \text{recall$_i$}}{\text{precision$_i$} + \text{recall$_i$}}.
\end{eqnarray}
Finally, the macro-averaged $F_1$ score is computed as the arithmetic mean of the $F_1$ scores across all $n$ classes: 
\begin{eqnarray}
\text{macro $F_{1}$} = \frac{1}{n}\sum_{i=1}^{n} F_{1i}.
\end{eqnarray}

\section{Results}
Motivated by the question of whether DINO-pretrained Vision Transformers, trained on either natural or domain-specific images, can generalize to other microscopy domains, we evaluated their performance on the downstream task of protein localization. Specifically, we compared how different pretrained embeddings perform when used to train a supervised classification head on OpenCell data.

Table \ref{tab:results} summarizes the results, detailing the pretrained weights, ViT architecture, embedding approach, number of DINO pretraining epochs, and the resulting performance. We report the mean macro~$F_1$ scores on the test sets. A first observation is that embeddings obtained from the DINO model pretrained on the microscopy-specific HPA dataset achieved the best test performance, reaching a mean macro~$F_1$ score of 0.8221~($\pm$0.0062), when using the channel mapping approach. Interestingly, the same pretrained weights yielded the lowest performance among all models when paired with the channel replication approach, dropping to 0.7681~($\pm$0.0140). In contrast, embeddings derived from the model pretrained on ImageNet-1k using channel mapping and the model trained directly on OpenCell data, produced comparable results, with $F_1$ scores of 0.8057~($\pm$0.0090) and 0.7918~($\pm$0.0096), respectively.

\begin{figure}
    \centering
    \includegraphics[width=0.75\linewidth]{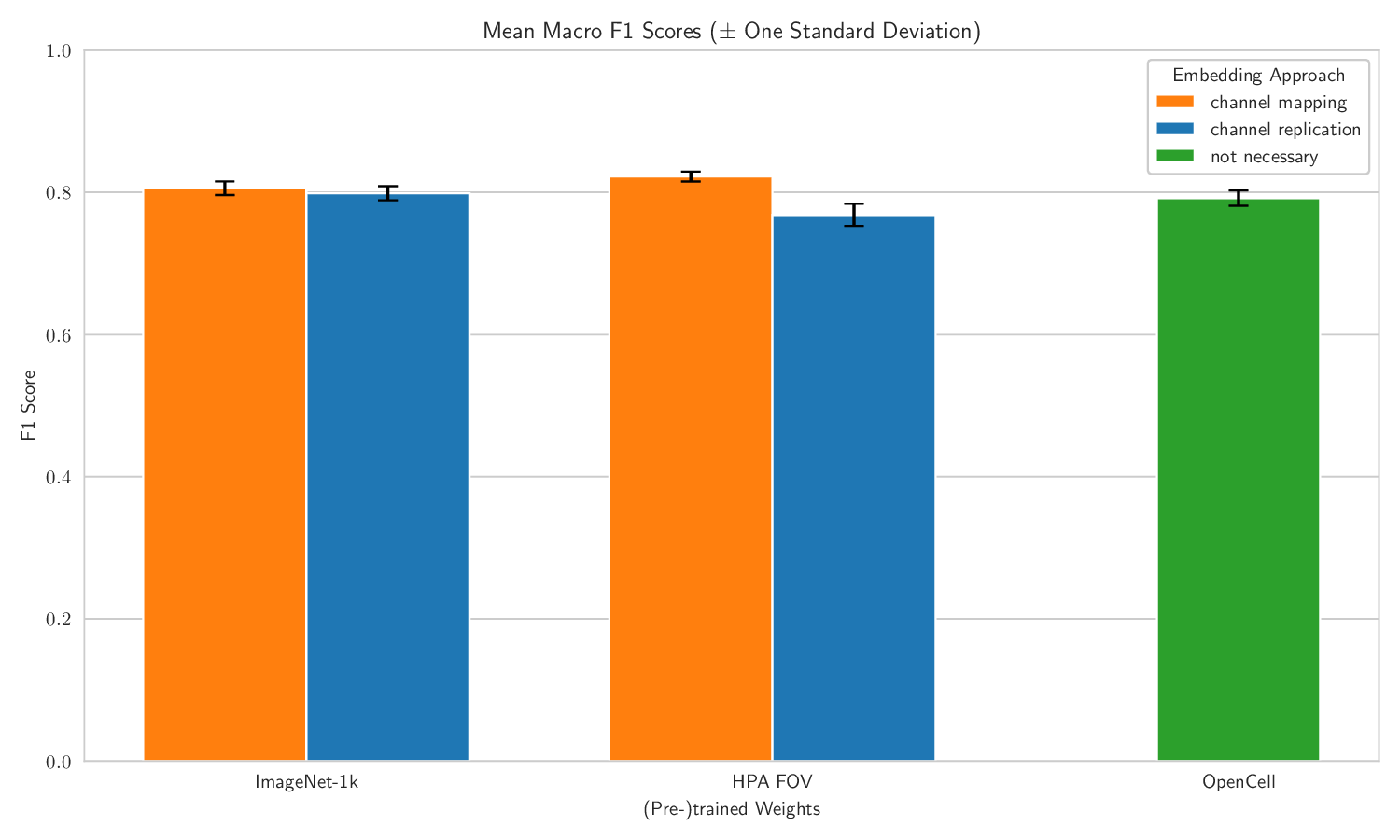}
    \caption{Comparison of mean macro F1 scores across different model configurations, evaluated over 5-fold cross-validation. Bars represent the mean F1 score, and error bars indicate $\pm$ one standard deviation across folds. Model variants are grouped by pretraining datasets (HPA FOV, ImageNet-1k, OpenCell) and embedding approach (channel replication vs. channel mapping).}
    \label{fig:std_results}
\end{figure}

While Table~\ref{tab:results} summarizes the minor differences between the (pre)trained DINO models, Figure~\ref{fig:std_results} illustrates their overall comparable performance by displaying error bars and standard deviations. 

\begin{figure}
    \centering
    \includegraphics[width=\linewidth]{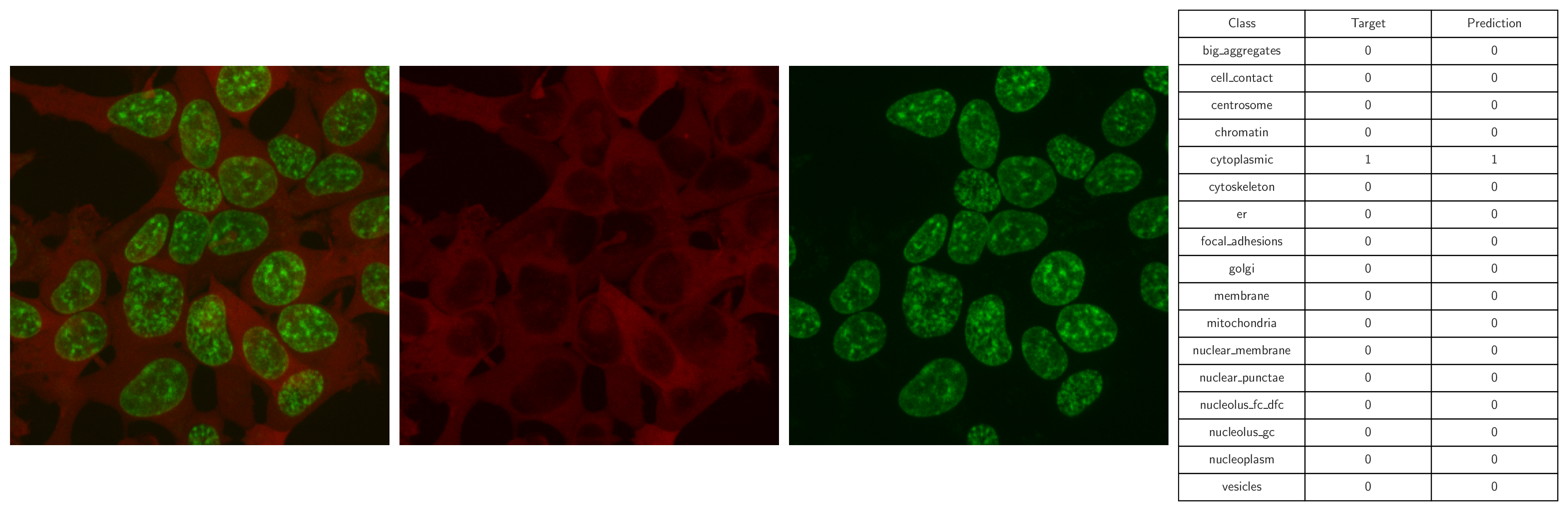}
    \includegraphics[width=\linewidth]{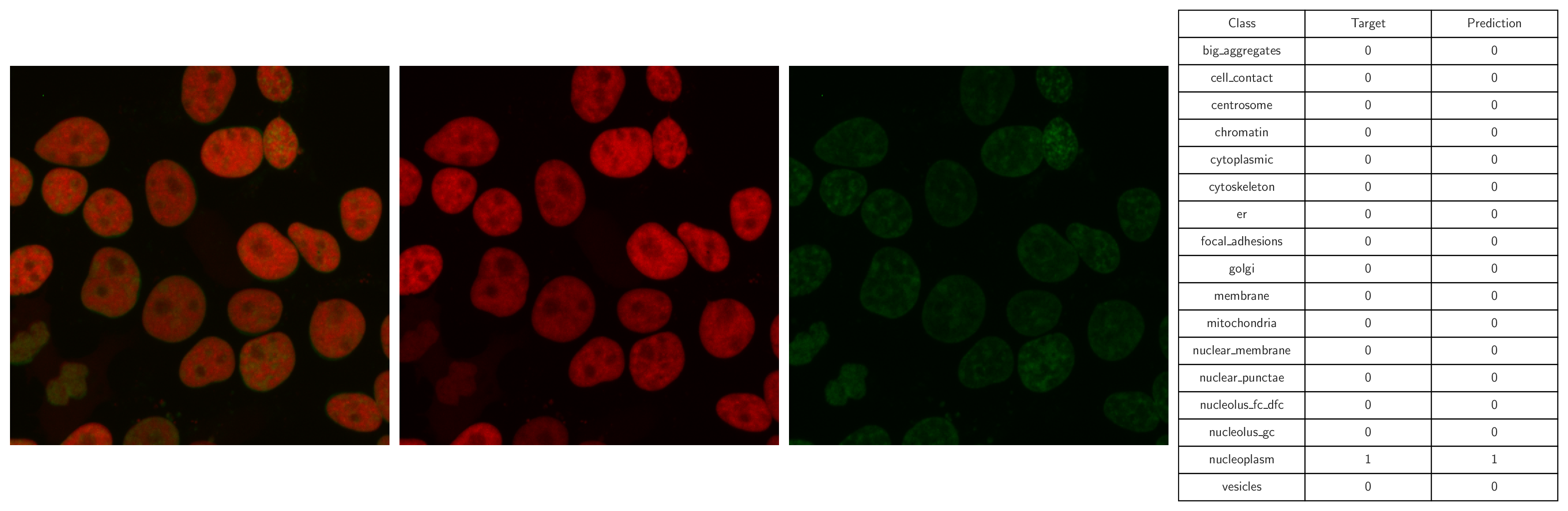}
    \caption{Visualization of the downstream analysis for two OpenCell samples, illustrating the predicted protein localization. The composite image (left) shows merged channels for protein (red) and nucleus (green). Individual channels are displayed separately (middle panels) alongside a classification table (right) with ground truth and predictions obtained by the best-performing fold of the 5-fold cross-validation using DINO pretrained on HPA FOV data with the channel mapping strategy.  For the protein's subcellular localization, the model accurately predicted both the cytoplasmic and nucleoplasm compartments.}
    \label{fig:example_predictions}
\end{figure}

From our cross-validation results, we selected the best-performing fold using DINO pretrained on HPA FOV data with channel mapping and visualized two example images. The composite views (Figure~\ref{fig:example_predictions}) reveal where protein (red) and nuclear (green) signals co-localize, while the separate channel displays show their unique distributions. Each image is accompanied by a classification table listing the predicted subcellular compartments alongside their target labels and models prediction.

    \begin{table}
    \caption{\label{tab:results}Model performance evaluation (mean macro F$_1$ $\pm$ one standard deviation (std)) for different pretraining configurations.}
    \begin{ruledtabular}
    \begin{tabular}{cccccc}
    (Pre)trained Weights & Model & Embedding Approach & DINO Epochs & Mean Macro $F_1$\\
    \hline
    ImageNet-1k & vit\_base/8 & channel mapping & 300 & 0.8057 ($\pm$ 0.0090) \\
    ImageNet-1k & vit\_base/8 & channel replication & 300 & 0.7986 ($\pm$ 0.0089)\\
    HPA FOV & vit\_base/8 & channel mapping & 100 & 0.8221 ($\pm$ 0.0062)\\
    HPA FOV & vit\_base/8 & channel replication & 100  & 0.7681 ($\pm$0.0140)\\
    OpenCell & vit\_base/8 & - & 300  & 0.7918 ($\pm$ 0.0096) \\
    \end{tabular}
    \end{ruledtabular}
    \end{table}

\section{Discussion}
This work focused on the investigation of different embedding methods and pretraining strategies on generalization and downstream performance in microscopy image analysis.

In sum, all considered embedding and pretraining approaches yielded F1 scores of roughly about 0.8. The best-performing model, which integrates a DINO pretrained on the HPA dataset, seems to benefits from both the microscopy-specific nature of the data and the large number of data points available in HPA. In comparison, the OpenCell dataset is approximately 38 times smaller than HPA. The visualization of image data using DINO pretrained on HPA FOV data demonstrates the model's ability to accurately predict subcellular localization, as seen in Figure~\ref{fig:example_predictions} for the correct classification of cytoplasmic and nucleoplasm compartments in the given images. 

We hypothesize that the DINO model pretrained on HPA performs slightly better on the OpenCell dataset than the model pretrained on ImageNet-1k due to the domain shift between microscopy images and natural images. Nevertheless, the ImageNet-1k-pretrained model remains competitive with the model trained from scratch on OpenCell, suggesting that the OpenCell dataset may not be large enough to fully leverage domain-specific information. Additionally, the ImageNet-1k dataset contains 1.2 million images, which likely contributes to the robustness of the pretrained model.

Pretrained models based on HPA and ImageNet-1k performed at least as well on HPA-related downstream tasks as models trained specifically on the smaller OpenCell dataset. This suggests that pretrained models generalize well across microscopy tasks, even when the target dataset is smaller or task-specific.

Regarding embedding methods, early fusion using channel mapping tended to perform slightly better than channel replication, as it allows the encoder to integrate spatial information across channels more effectively. In contrast, late fusion with channel replication performed slightly worse, likely due to the added noise introduced when concatenating multiple embedding vectors.

Across all settings, the difference between training performance ($\geq 0.94$ macro F1 points in all scenarios) and test performance was approximately 0.2 macro~$F_1$ points, suggesting some degree of overfitting. Nevertheless, the models generalized reasonably well to unseen data, as reflected in consistently strong test performance. 

As shown in Figure~\ref{fig:std_results}, all experiments demonstrated comparable performance, indicating that pretrained models generalize effectively across different (microscopy) datasets and embedding approaches. Our results highlight that pretrained DINO representations, even when trained on natural or unrelated cell images, encode robust morphological features useful for downstream tasks like protein localization. This reduces the need for task-specific pretraining or fine-tuning. Performance differences across embedding strategies emphasize the importance of input channel handling. Limitations include the static classifier architecture and limited exploration of multi-task or semi-supervised fine-tuning.

\section{Conclusion and Future Work}
This study demonstrates that ViT-based self-supervised models pretrained on HPA or ImageNet-1k generalize well to a distinct dataset (OpenCell) for the task of protein localization classification. These findings pave the way for broader adoption of pretrained models in microscopy and highlight the potential of transfer learning in bioimage analysis. Future work should explore domain adaptation strategies and end-to-end fine-tuning on weakly labeled or small-scale datasets to further enhance performance and robustness.

% \section{Conclusion}

% In this section we welcome you to include a summary of the end results of your research.

\begin{acknowledgments}
The authors acknowledge funding by the DFG Project INST 168/4-1.
\end{acknowledgments}

%\nocite{*}
% \bibliography{sn-bibliography}   % disable BibTeX on arXiv

\begin{thebibliography}{50}%
\makeatletter
\providecommand \@ifxundefined [1]{%
 \@ifx{#1\undefined}
}%
\providecommand \@ifnum [1]{%
 \ifnum #1\expandafter \@firstoftwo
 \else \expandafter \@secondoftwo
 \fi
}%
\providecommand \@ifx [1]{%
 \ifx #1\expandafter \@firstoftwo
 \else \expandafter \@secondoftwo
 \fi
}%
\providecommand \natexlab [1]{#1}%
\providecommand \enquote  [1]{``#1''}%
\providecommand \bibnamefont  [1]{#1}%
\providecommand \bibfnamefont [1]{#1}%
\providecommand \citenamefont [1]{#1}%
\providecommand \href@noop [0]{\@secondoftwo}%
\providecommand \href [0]{\begingroup \@sanitize@url \@href}%
\providecommand \@href[1]{\@@startlink{#1}\@@href}%
\providecommand \@@href[1]{\endgroup#1\@@endlink}%
\providecommand \@sanitize@url [0]{\catcode `\\12\catcode `\$12\catcode
  `\&12\catcode `\#12\catcode `\^12\catcode `\_12\catcode `\%12\relax}%
\providecommand \@@startlink[1]{}%
\providecommand \@@endlink[0]{}%
\providecommand \url  [0]{\begingroup\@sanitize@url \@url }%
\providecommand \@url [1]{\endgroup\@href {#1}{\urlprefix }}%
\providecommand \urlprefix  [0]{URL }%
\providecommand \Eprint [0]{\href }%
\providecommand \doibase [0]{http://dx.doi.org/}%
\providecommand \selectlanguage [0]{\@gobble}%
\providecommand \bibinfo  [0]{\@secondoftwo}%
\providecommand \bibfield  [0]{\@secondoftwo}%
\providecommand \translation [1]{[#1]}%
\providecommand \BibitemOpen [0]{}%
\providecommand \bibitemStop [0]{}%
\providecommand \bibitemNoStop [0]{.\EOS\space}%
\providecommand \EOS [0]{\spacefactor3000\relax}%
\providecommand \BibitemShut  [1]{\csname bibitem#1\endcsname}%
\let\auto@bib@innerbib\@empty
%</preamble>
\bibitem [{\citenamefont {Jia}\ \emph {et~al.}(2022)\citenamefont {Jia},
  \citenamefont {Chu}, \citenamefont {Jin}, \citenamefont {Long},\ and\
  \citenamefont {Zhu}}]{jia_high-throughput_2022}%
  \BibitemOpen
  \bibfield  {author} {\bibinfo {author} {\bibfnamefont {Q.}~\bibnamefont
  {Jia}}, \bibinfo {author} {\bibfnamefont {H.}~\bibnamefont {Chu}}, \bibinfo
  {author} {\bibfnamefont {Z.}~\bibnamefont {Jin}}, \bibinfo {author}
  {\bibfnamefont {H.}~\bibnamefont {Long}}, \ and\ \bibinfo {author}
  {\bibfnamefont {B.}~\bibnamefont {Zhu}},\ }\bibfield  {title}
  {{\selectlanguage{english}\enquote {\bibinfo {title} {High-throughput single-cell
  sequencing in cancer research},}\ }}\href {\doibase
  10.1038/s41392-022-00990-4} {\bibfield  {journal} {\bibinfo  {journal}
  {Signal Transduction and Targeted Therapy}\ }\textbf {\bibinfo {volume}
  {7}},\ \bibinfo {pages} {1--20} (\bibinfo {year} {2022})},\ \bibinfo {note}
  {publisher: Nature Publishing Group}\BibitemShut {NoStop}%
\bibitem [{\citenamefont {Navin}\ and\ \citenamefont
  {Hicks}(2011)}]{navin_future_2011}%
  \BibitemOpen
  \bibfield  {author} {\bibinfo {author} {\bibfnamefont {N.}~\bibnamefont
  {Navin}}\ and\ \bibinfo {author} {\bibfnamefont {J.}~\bibnamefont {Hicks}},\
  }\bibfield  {title} {\enquote {\bibinfo {title} {Future medical applications
  of single-cell sequencing in cancer},}\ }\href {\doibase 10.1186/gm247}
  {\bibfield  {journal} {\bibinfo  {journal} {Genome Medicine}\ }\textbf
  {\bibinfo {volume} {3}},\ \bibinfo {pages} {31} (\bibinfo {year}
  {2011})}\BibitemShut {NoStop}%
\bibitem [{\citenamefont {Buenrostro}\ \emph {et~al.}(2015)\citenamefont
  {Buenrostro}, \citenamefont {Wu}, \citenamefont {Litzenburger}, \citenamefont
  {Ruff}, \citenamefont {Gonzales}, \citenamefont {Snyder}, \citenamefont
  {Chang},\ and\ \citenamefont {Greenleaf}}]{buenrostro_single-cell_2015}%
  \BibitemOpen
  \bibfield  {author} {\bibinfo {author} {\bibfnamefont {J.~D.}\ \bibnamefont
  {Buenrostro}}, \bibinfo {author} {\bibfnamefont {B.}~\bibnamefont {Wu}},
  \bibinfo {author} {\bibfnamefont {U.~M.}\ \bibnamefont {Litzenburger}},
  \bibinfo {author} {\bibfnamefont {D.}~\bibnamefont {Ruff}}, \bibinfo {author}
  {\bibfnamefont {M.~L.}\ \bibnamefont {Gonzales}}, \bibinfo {author}
  {\bibfnamefont {M.~P.}\ \bibnamefont {Snyder}}, \bibinfo {author}
  {\bibfnamefont {H.~Y.}\ \bibnamefont {Chang}}, \ and\ \bibinfo {author}
  {\bibfnamefont {W.~J.}\ \bibnamefont {Greenleaf}},\ }\bibfield  {title}
  {{\selectlanguage{english}\enquote {\bibinfo {title} {Single-cell chromatin
  accessibility reveals principles of regulatory variation},}\ }}\href
  {\doibase 10.1038/nature14590} {\bibfield  {journal} {\bibinfo  {journal}
  {Nature}\ }\textbf {\bibinfo {volume} {523}},\ \bibinfo {pages} {486--490}
  (\bibinfo {year} {2015})}\BibitemShut {NoStop}%
\bibitem [{\citenamefont {Bandura}\ \emph {et~al.}(2009)\citenamefont
  {Bandura}, \citenamefont {Baranov}, \citenamefont {Ornatsky}, \citenamefont
  {Antonov}, \citenamefont {Kinach}, \citenamefont {Lou}, \citenamefont
  {Pavlov}, \citenamefont {Vorobiev}, \citenamefont {Dick},\ and\ \citenamefont
  {Tanner}}]{bandura_mass_2009}%
  \BibitemOpen
  \bibfield  {author} {\bibinfo {author} {\bibfnamefont {D.~R.}\ \bibnamefont
  {Bandura}}, \bibinfo {author} {\bibfnamefont {V.~I.}\ \bibnamefont
  {Baranov}}, \bibinfo {author} {\bibfnamefont {O.~I.}\ \bibnamefont
  {Ornatsky}}, \bibinfo {author} {\bibfnamefont {A.}~\bibnamefont {Antonov}},
  \bibinfo {author} {\bibfnamefont {R.}~\bibnamefont {Kinach}}, \bibinfo
  {author} {\bibfnamefont {X.}~\bibnamefont {Lou}}, \bibinfo {author}
  {\bibfnamefont {S.}~\bibnamefont {Pavlov}}, \bibinfo {author} {\bibfnamefont
  {S.}~\bibnamefont {Vorobiev}}, \bibinfo {author} {\bibfnamefont {J.~E.}\
  \bibnamefont {Dick}}, \ and\ \bibinfo {author} {\bibfnamefont {S.~D.}\
  \bibnamefont {Tanner}},\ }\bibfield  {title} {\enquote {\bibinfo {title}
  {Mass {Cytometry}: {Technique} for {Real} {Time} {Single} {Cell}
  {Multitarget} {Immunoassay} {Based} on {Inductively} {Coupled} {Plasma}
  {Time}-of-{Flight} {Mass} {Spectrometry}},}\ }\href {\doibase
  10.1021/ac901049w} {\bibfield  {journal} {\bibinfo  {journal} {Analytical
  Chemistry}\ }\textbf {\bibinfo {volume} {81}},\ \bibinfo {pages} {6813--6822}
  (\bibinfo {year} {2009})},\ \bibinfo {note} {publisher: American Chemical
  Society}\BibitemShut {NoStop}%
\bibitem [{\citenamefont {Wang}\ \emph {et~al.}(2020)\citenamefont {Wang},
  \citenamefont {Mo}, \citenamefont {Li}, \citenamefont {He},\ and\
  \citenamefont {Yang}}]{wang_single-cell_2020}%
  \BibitemOpen
  \bibfield  {author} {\bibinfo {author} {\bibfnamefont {L.}~\bibnamefont
  {Wang}}, \bibinfo {author} {\bibfnamefont {S.}~\bibnamefont {Mo}}, \bibinfo
  {author} {\bibfnamefont {X.}~\bibnamefont {Li}}, \bibinfo {author}
  {\bibfnamefont {Y.}~\bibnamefont {He}}, \ and\ \bibinfo {author}
  {\bibfnamefont {J.}~\bibnamefont {Yang}},\ }\bibfield  {title}
  {{\selectlanguage{english}\enquote {\bibinfo {title} {Single-cell {RNA}-seq
  reveals the immune escape and drug resistance mechanisms of mantle cell
  lymphoma},}\ }}\href {\doibase 10.20892/j.issn.2095-3941.2020.0073}
  {\bibfield  {journal} {\bibinfo  {journal} {Cancer Biology and Medicine}\
  }\textbf {\bibinfo {volume} {17}},\ \bibinfo {pages} {726--739} (\bibinfo
  {year} {2020})}\BibitemShut {NoStop}%
\bibitem [{\citenamefont {Heumos}\ \emph {et~al.}(2023)\citenamefont {Heumos},
  \citenamefont {Schaar}, \citenamefont {Lance}, \citenamefont {Litinetskaya},
  \citenamefont {Drost}, \citenamefont {Zappia}, \citenamefont {Lücken},
  \citenamefont {Strobl}, \citenamefont {Henao}, \citenamefont {Curion},
  \citenamefont {Schiller},\ and\ \citenamefont {Theis}}]{heumos_best_2023}%
  \BibitemOpen
  \bibfield  {author} {\bibinfo {author} {\bibfnamefont {L.}~\bibnamefont
  {Heumos}}, \bibinfo {author} {\bibfnamefont {A.~C.}\ \bibnamefont {Schaar}},
  \bibinfo {author} {\bibfnamefont {C.}~\bibnamefont {Lance}}, \bibinfo
  {author} {\bibfnamefont {A.}~\bibnamefont {Litinetskaya}}, \bibinfo {author}
  {\bibfnamefont {F.}~\bibnamefont {Drost}}, \bibinfo {author} {\bibfnamefont
  {L.}~\bibnamefont {Zappia}}, \bibinfo {author} {\bibfnamefont {M.~D.}\
  \bibnamefont {Lücken}}, \bibinfo {author} {\bibfnamefont {D.~C.}\
  \bibnamefont {Strobl}}, \bibinfo {author} {\bibfnamefont {J.}~\bibnamefont
  {Henao}}, \bibinfo {author} {\bibfnamefont {F.}~\bibnamefont {Curion}},
  \bibinfo {author} {\bibfnamefont {H.~B.}\ \bibnamefont {Schiller}}, \ and\
  \bibinfo {author} {\bibfnamefont {F.~J.}\ \bibnamefont {Theis}},\ }\bibfield
  {title} {{\selectlanguage{english}\enquote {\bibinfo {title} {Best practices for
  single-cell analysis across modalities},}\ }}\href {\doibase
  10.1038/s41576-023-00586-w} {\bibfield  {journal} {\bibinfo  {journal}
  {Nature Reviews Genetics}\ }\textbf {\bibinfo {volume} {24}},\ \bibinfo
  {pages} {550--572} (\bibinfo {year} {2023})},\ \bibinfo {note} {publisher:
  Nature Publishing Group}\BibitemShut {NoStop}%
\bibitem [{\citenamefont {Mattiazzi~Usaj}\ \emph {et~al.}(2016)\citenamefont
  {Mattiazzi~Usaj}, \citenamefont {Styles}, \citenamefont {Verster},
  \citenamefont {Friesen}, \citenamefont {Boone},\ and\ \citenamefont
  {Andrews}}]{mattiazzi_usaj_high-content_2016}%
  \BibitemOpen
  \bibfield  {author} {\bibinfo {author} {\bibfnamefont {M.}~\bibnamefont
  {Mattiazzi~Usaj}}, \bibinfo {author} {\bibfnamefont {E.~B.}\ \bibnamefont
  {Styles}}, \bibinfo {author} {\bibfnamefont {A.~J.}\ \bibnamefont {Verster}},
  \bibinfo {author} {\bibfnamefont {H.}~\bibnamefont {Friesen}}, \bibinfo
  {author} {\bibfnamefont {C.}~\bibnamefont {Boone}}, \ and\ \bibinfo {author}
  {\bibfnamefont {B.~J.}\ \bibnamefont {Andrews}},\ }\bibfield  {title}
  {\enquote {\bibinfo {title} {High-{Content} {Screening} for {Quantitative}
  {Cell} {Biology}},}\ }\href {\doibase 10.1016/j.tcb.2016.03.008} {\bibfield
  {journal} {\bibinfo  {journal} {Trends in Cell Biology}\ }\textbf {\bibinfo
  {volume} {26}},\ \bibinfo {pages} {598--611} (\bibinfo {year}
  {2016})}\BibitemShut {NoStop}%
\bibitem [{\citenamefont {Grys}\ \emph {et~al.}(2016)\citenamefont {Grys},
  \citenamefont {Lo}, \citenamefont {Sahin}, \citenamefont {Kraus},
  \citenamefont {Morris}, \citenamefont {Boone},\ and\ \citenamefont
  {Andrews}}]{grys_machine_2016}%
  \BibitemOpen
  \bibfield  {author} {\bibinfo {author} {\bibfnamefont {B.~T.}\ \bibnamefont
  {Grys}}, \bibinfo {author} {\bibfnamefont {D.~S.}\ \bibnamefont {Lo}},
  \bibinfo {author} {\bibfnamefont {N.}~\bibnamefont {Sahin}}, \bibinfo
  {author} {\bibfnamefont {O.~Z.}\ \bibnamefont {Kraus}}, \bibinfo {author}
  {\bibfnamefont {Q.}~\bibnamefont {Morris}}, \bibinfo {author} {\bibfnamefont
  {C.}~\bibnamefont {Boone}}, \ and\ \bibinfo {author} {\bibfnamefont {B.~J.}\
  \bibnamefont {Andrews}},\ }\bibfield  {title} {\enquote {\bibinfo {title}
  {Machine learning and computer vision approaches for phenotypic profiling},}\
  }\href {\doibase 10.1083/jcb.201610026} {\bibfield  {journal} {\bibinfo
  {journal} {Journal of Cell Biology}\ }\textbf {\bibinfo {volume} {216}},\
  \bibinfo {pages} {65--71} (\bibinfo {year} {2016})}\BibitemShut {NoStop}%
\bibitem [{\citenamefont {Scheeder}, \citenamefont {Heigwer},\ and\
  \citenamefont {Boutros}(2018)}]{scheeder_machine_2018}%
  \BibitemOpen
  \bibfield  {author} {\bibinfo {author} {\bibfnamefont {C.}~\bibnamefont
  {Scheeder}}, \bibinfo {author} {\bibfnamefont {F.}~\bibnamefont {Heigwer}}, \
  and\ \bibinfo {author} {\bibfnamefont {M.}~\bibnamefont {Boutros}},\
  }\bibfield  {title} {\enquote {\bibinfo {title} {Machine learning and
  image-based profiling in drug discovery},}\ }\href {\doibase
  10.1016/j.coisb.2018.05.004} {\bibfield  {journal} {\bibinfo  {journal}
  {Current Opinion in Systems Biology}\ }\bibinfo {series} {Pharmacology and
  drug discovery},\ \textbf {\bibinfo {volume} {10}},\ \bibinfo {pages}
  {43--52} (\bibinfo {year} {2018})}\BibitemShut {NoStop}%
\bibitem [{\citenamefont {Reicher}\ \emph {et~al.}(2024)\citenamefont
  {Reicher}, \citenamefont {Reiniš}, \citenamefont {Ciobanu}, \citenamefont
  {Růžička}, \citenamefont {Malik}, \citenamefont {Siklos}, \citenamefont
  {Kartysh}, \citenamefont {Tomek}, \citenamefont {Koren}, \citenamefont
  {Rendeiro},\ and\ \citenamefont {Kubicek}}]{reicher_pooled_2024}%
  \BibitemOpen
  \bibfield  {author} {\bibinfo {author} {\bibfnamefont {A.}~\bibnamefont
  {Reicher}}, \bibinfo {author} {\bibfnamefont {J.}~\bibnamefont {Reiniš}},
  \bibinfo {author} {\bibfnamefont {M.}~\bibnamefont {Ciobanu}}, \bibinfo
  {author} {\bibfnamefont {P.}~\bibnamefont {Růžička}}, \bibinfo {author}
  {\bibfnamefont {M.}~\bibnamefont {Malik}}, \bibinfo {author} {\bibfnamefont
  {M.}~\bibnamefont {Siklos}}, \bibinfo {author} {\bibfnamefont
  {V.}~\bibnamefont {Kartysh}}, \bibinfo {author} {\bibfnamefont
  {T.}~\bibnamefont {Tomek}}, \bibinfo {author} {\bibfnamefont
  {A.}~\bibnamefont {Koren}}, \bibinfo {author} {\bibfnamefont {A.~F.}\
  \bibnamefont {Rendeiro}}, \ and\ \bibinfo {author} {\bibfnamefont
  {S.}~\bibnamefont {Kubicek}},\ }\bibfield  {title} {{\selectlanguage{english}\enquote {\bibinfo {title} {Pooled multicolour tagging for visualizing
  subcellular protein dynamics},}\ }}\href {\doibase
  10.1038/s41556-024-01407-w} {\bibfield  {journal} {\bibinfo  {journal}
  {Nature Cell Biology}\ }\textbf {\bibinfo {volume} {26}},\ \bibinfo {pages}
  {745--756} (\bibinfo {year} {2024})},\ \bibinfo {note} {publisher: Nature
  Publishing Group}\BibitemShut {NoStop}%
\bibitem [{\citenamefont {Vincent}\ \emph {et~al.}(2022)\citenamefont
  {Vincent}, \citenamefont {Nueda}, \citenamefont {Lee}, \citenamefont
  {Schenone}, \citenamefont {Prunotto},\ and\ \citenamefont
  {Mercola}}]{vincent_phenotypic_2022}%
  \BibitemOpen
  \bibfield  {author} {\bibinfo {author} {\bibfnamefont {F.}~\bibnamefont
  {Vincent}}, \bibinfo {author} {\bibfnamefont {A.}~\bibnamefont {Nueda}},
  \bibinfo {author} {\bibfnamefont {J.}~\bibnamefont {Lee}}, \bibinfo {author}
  {\bibfnamefont {M.}~\bibnamefont {Schenone}}, \bibinfo {author}
  {\bibfnamefont {M.}~\bibnamefont {Prunotto}}, \ and\ \bibinfo {author}
  {\bibfnamefont {M.}~\bibnamefont {Mercola}},\ }\bibfield  {title}
  {{\selectlanguage{english}\enquote {\bibinfo {title} {Phenotypic drug discovery:
  recent successes, lessons learned and new directions},}\ }}\href {\doibase
  10.1038/s41573-022-00472-w} {\bibfield  {journal} {\bibinfo  {journal}
  {Nature Reviews Drug Discovery}\ }\textbf {\bibinfo {volume} {21}},\ \bibinfo
  {pages} {899--914} (\bibinfo {year} {2022})},\ \bibinfo {note} {publisher:
  Nature Publishing Group}\BibitemShut {NoStop}%
\bibitem [{\citenamefont {Boyd}, \citenamefont {Fennell},\ and\ \citenamefont
  {Carpenter}(2020)}]{boyd_harnessing_2020}%
  \BibitemOpen
  \bibfield  {author} {\bibinfo {author} {\bibfnamefont {J.}~\bibnamefont
  {Boyd}}, \bibinfo {author} {\bibfnamefont {M.}~\bibnamefont {Fennell}}, \
  and\ \bibinfo {author} {\bibfnamefont {A.}~\bibnamefont {Carpenter}},\
  }\bibfield  {title} {\enquote {\bibinfo {title} {Harnessing the power of
  microscopy images to accelerate drug discovery: what are the
  possibilities?}}\ }\href {\doibase 10.1080/17460441.2020.1743675} {\bibfield
  {journal} {\bibinfo  {journal} {Expert Opinion on Drug Discovery}\ }\textbf
  {\bibinfo {volume} {15}},\ \bibinfo {pages} {639--642} (\bibinfo {year}
  {2020})},\ \bibinfo {note} {publisher: Taylor \& Francis \_eprint:
  https://doi.org/10.1080/17460441.2020.1743675}\BibitemShut {NoStop}%
\bibitem [{\citenamefont {LeCun}, \citenamefont {Bengio},\ and\ \citenamefont
  {Hinton}(2015)}]{lecun_deep_2015}%
  \BibitemOpen
  \bibfield  {author} {\bibinfo {author} {\bibfnamefont {Y.}~\bibnamefont
  {LeCun}}, \bibinfo {author} {\bibfnamefont {Y.}~\bibnamefont {Bengio}}, \
  and\ \bibinfo {author} {\bibfnamefont {G.}~\bibnamefont {Hinton}},\
  }\bibfield  {title} {{\selectlanguage{english}\enquote {\bibinfo {title} {Deep
  learning},}\ }}\href {\doibase 10.1038/nature14539} {\bibfield  {journal}
  {\bibinfo  {journal} {Nature}\ }\textbf {\bibinfo {volume} {521}},\ \bibinfo
  {pages} {436--444} (\bibinfo {year} {2015})},\ \bibinfo {note} {publisher:
  Nature Publishing Group}\BibitemShut {NoStop}%
\bibitem [{\citenamefont {Premkumar}\ \emph {et~al.}(2024)\citenamefont
  {Premkumar}, \citenamefont {Srinivasan}, \citenamefont {Harini~Devi},
  \citenamefont {M}, \citenamefont {E}, \citenamefont {Jadhav}, \citenamefont
  {Futane},\ and\ \citenamefont {Narayanamurthy}}]{premkumar_single-cell_2024}%
  \BibitemOpen
  \bibfield  {author} {\bibinfo {author} {\bibfnamefont {R.}~\bibnamefont
  {Premkumar}}, \bibinfo {author} {\bibfnamefont {A.}~\bibnamefont
  {Srinivasan}}, \bibinfo {author} {\bibfnamefont {K.~G.}\ \bibnamefont
  {Harini~Devi}}, \bibinfo {author} {\bibfnamefont {D.}~\bibnamefont {M}},
  \bibinfo {author} {\bibfnamefont {G.}~\bibnamefont {E}}, \bibinfo {author}
  {\bibfnamefont {P.}~\bibnamefont {Jadhav}}, \bibinfo {author} {\bibfnamefont
  {A.}~\bibnamefont {Futane}}, \ and\ \bibinfo {author} {\bibfnamefont
  {V.}~\bibnamefont {Narayanamurthy}},\ }\bibfield  {title} {\enquote {\bibinfo
  {title} {Single-cell classification, analysis, and its application using deep
  learning techniques},}\ }\href {\doibase 10.1016/j.biosystems.2024.105142}
  {\bibfield  {journal} {\bibinfo  {journal} {BioSystems}\ }\textbf {\bibinfo
  {volume} {237}},\ \bibinfo {pages} {105142} (\bibinfo {year}
  {2024})}\BibitemShut {NoStop}%
\bibitem [{\citenamefont {Kim}\ \emph {et~al.}(2025)\citenamefont {Kim},
  \citenamefont {Adaloglou}, \citenamefont {Osterland}, \citenamefont
  {Morelli}, \citenamefont {Halawa}, \citenamefont {König}, \citenamefont
  {Gnutt},\ and\ \citenamefont {Marin~Zapata}}]{kim_self-supervision_2025}%
  \BibitemOpen
  \bibfield  {author} {\bibinfo {author} {\bibfnamefont {V.}~\bibnamefont
  {Kim}}, \bibinfo {author} {\bibfnamefont {N.}~\bibnamefont {Adaloglou}},
  \bibinfo {author} {\bibfnamefont {M.}~\bibnamefont {Osterland}}, \bibinfo
  {author} {\bibfnamefont {F.~M.}\ \bibnamefont {Morelli}}, \bibinfo {author}
  {\bibfnamefont {M.}~\bibnamefont {Halawa}}, \bibinfo {author} {\bibfnamefont
  {T.}~\bibnamefont {König}}, \bibinfo {author} {\bibfnamefont
  {D.}~\bibnamefont {Gnutt}}, \ and\ \bibinfo {author} {\bibfnamefont {P.~A.}\
  \bibnamefont {Marin~Zapata}},\ }\bibfield  {title} {{\selectlanguage{english}\enquote {\bibinfo {title} {Self-supervision advances morphological
  profiling by unlocking powerful image representations},}\ }}\href {\doibase
  10.1038/s41598-025-88825-4} {\bibfield  {journal} {\bibinfo  {journal}
  {Scientific Reports}\ }\textbf {\bibinfo {volume} {15}},\ \bibinfo {pages}
  {4876} (\bibinfo {year} {2025})},\ \bibinfo {note} {publisher: Nature
  Publishing Group}\BibitemShut {NoStop}%
\bibitem [{\citenamefont {Pham}, \citenamefont {Caicedo},\ and\ \citenamefont
  {Plummer}(2025)}]{pham_cha-maevit_2025}%
  \BibitemOpen
  \bibfield  {author} {\bibinfo {author} {\bibfnamefont {C.}~\bibnamefont
  {Pham}}, \bibinfo {author} {\bibfnamefont {J.~C.}\ \bibnamefont {Caicedo}}, \
  and\ \bibinfo {author} {\bibfnamefont {B.~A.}\ \bibnamefont {Plummer}},\
  }\href {\doibase 10.48550/arXiv.2503.19331} {\enquote {\bibinfo {title}
  {{ChA}-{MAEViT}: {Unifying} {Channel}-{Aware} {Masked} {Autoencoders} and
  {Multi}-{Channel} {Vision} {Transformers} for {Improved} {Cross}-{Channel}
  {Learning}},}\ } (\bibinfo {year} {2025}),\ \bibinfo {note} {arXiv:2503.19331
  [cs]}\BibitemShut {NoStop}%
\bibitem [{\citenamefont {Kraus}\ \emph {et~al.}(2024)\citenamefont {Kraus},
  \citenamefont {Kenyon-Dean}, \citenamefont {Saberian}, \citenamefont
  {Fallah}, \citenamefont {McLean}, \citenamefont {Leung}, \citenamefont
  {Sharma}, \citenamefont {Khan}, \citenamefont {Balakrishnan}, \citenamefont
  {Celik}, \citenamefont {Beaini}, \citenamefont {Sypetkowski}, \citenamefont
  {Cheng}, \citenamefont {Morse}, \citenamefont {Makes}, \citenamefont
  {Mabey},\ and\ \citenamefont {Earnshaw}}]{kraus_masked_2024}%
  \BibitemOpen
  \bibfield  {author} {\bibinfo {author} {\bibfnamefont {O.}~\bibnamefont
  {Kraus}}, \bibinfo {author} {\bibfnamefont {K.}~\bibnamefont {Kenyon-Dean}},
  \bibinfo {author} {\bibfnamefont {S.}~\bibnamefont {Saberian}}, \bibinfo
  {author} {\bibfnamefont {M.}~\bibnamefont {Fallah}}, \bibinfo {author}
  {\bibfnamefont {P.}~\bibnamefont {McLean}}, \bibinfo {author} {\bibfnamefont
  {J.}~\bibnamefont {Leung}}, \bibinfo {author} {\bibfnamefont
  {V.}~\bibnamefont {Sharma}}, \bibinfo {author} {\bibfnamefont
  {A.}~\bibnamefont {Khan}}, \bibinfo {author} {\bibfnamefont {J.}~\bibnamefont
  {Balakrishnan}}, \bibinfo {author} {\bibfnamefont {S.}~\bibnamefont {Celik}},
  \bibinfo {author} {\bibfnamefont {D.}~\bibnamefont {Beaini}}, \bibinfo
  {author} {\bibfnamefont {M.}~\bibnamefont {Sypetkowski}}, \bibinfo {author}
  {\bibfnamefont {C.~V.}\ \bibnamefont {Cheng}}, \bibinfo {author}
  {\bibfnamefont {K.}~\bibnamefont {Morse}}, \bibinfo {author} {\bibfnamefont
  {M.}~\bibnamefont {Makes}}, \bibinfo {author} {\bibfnamefont
  {B.}~\bibnamefont {Mabey}}, \ and\ \bibinfo {author} {\bibfnamefont
  {B.}~\bibnamefont {Earnshaw}},\ }\href {\doibase 10.48550/arXiv.2404.10242}
  {\enquote {\bibinfo {title} {Masked {Autoencoders} for {Microscopy} are
  {Scalable} {Learners} of {Cellular} {Biology}},}\ } (\bibinfo {year}
  {2024}),\ \bibinfo {note} {arXiv:2404.10242 [cs]}\BibitemShut {NoStop}%
\bibitem [{\citenamefont {Doron}\ \emph {et~al.}(2023)\citenamefont {Doron},
  \citenamefont {Moutakanni}, \citenamefont {Chen}, \citenamefont {Moshkov},
  \citenamefont {Caron}, \citenamefont {Touvron}, \citenamefont {Bojanowski},
  \citenamefont {Pernice},\ and\ \citenamefont
  {Caicedo}}]{doron_unbiased_2023}%
  \BibitemOpen
  \bibfield  {author} {\bibinfo {author} {\bibfnamefont {M.}~\bibnamefont
  {Doron}}, \bibinfo {author} {\bibfnamefont {T.}~\bibnamefont {Moutakanni}},
  \bibinfo {author} {\bibfnamefont {Z.~S.}\ \bibnamefont {Chen}}, \bibinfo
  {author} {\bibfnamefont {N.}~\bibnamefont {Moshkov}}, \bibinfo {author}
  {\bibfnamefont {M.}~\bibnamefont {Caron}}, \bibinfo {author} {\bibfnamefont
  {H.}~\bibnamefont {Touvron}}, \bibinfo {author} {\bibfnamefont
  {P.}~\bibnamefont {Bojanowski}}, \bibinfo {author} {\bibfnamefont {W.~M.}\
  \bibnamefont {Pernice}}, \ and\ \bibinfo {author} {\bibfnamefont {J.~C.}\
  \bibnamefont {Caicedo}},\ }\href {\doibase 10.1101/2023.06.16.545359}
  {{\selectlanguage{english}\enquote {\bibinfo {title} {Unbiased single-cell
  morphology with self-supervised vision transformers},}\ }} (\bibinfo {year}
  {2023}),\ \bibinfo {note} {pages: 2023.06.16.545359 Section: New
  Results}\BibitemShut {NoStop}%
\bibitem [{\citenamefont {Gustafsdottir}\ \emph {et~al.}(2013)\citenamefont
  {Gustafsdottir}, \citenamefont {Ljosa}, \citenamefont {Sokolnicki},
  \citenamefont {Wilson}, \citenamefont {Walpita}, \citenamefont {Kemp},
  \citenamefont {Seiler}, \citenamefont {Carrel}, \citenamefont {Golub},
  \citenamefont {Schreiber}, \citenamefont {Clemons}, \citenamefont
  {Carpenter},\ and\ \citenamefont {Shamji}}]{gustafsdottir_multiplex_2013}%
  \BibitemOpen
  \bibfield  {author} {\bibinfo {author} {\bibfnamefont {S.~M.}\ \bibnamefont
  {Gustafsdottir}}, \bibinfo {author} {\bibfnamefont {V.}~\bibnamefont
  {Ljosa}}, \bibinfo {author} {\bibfnamefont {K.~L.}\ \bibnamefont
  {Sokolnicki}}, \bibinfo {author} {\bibfnamefont {J.~A.}\ \bibnamefont
  {Wilson}}, \bibinfo {author} {\bibfnamefont {D.}~\bibnamefont {Walpita}},
  \bibinfo {author} {\bibfnamefont {M.~M.}\ \bibnamefont {Kemp}}, \bibinfo
  {author} {\bibfnamefont {K.~P.}\ \bibnamefont {Seiler}}, \bibinfo {author}
  {\bibfnamefont {H.~A.}\ \bibnamefont {Carrel}}, \bibinfo {author}
  {\bibfnamefont {T.~R.}\ \bibnamefont {Golub}}, \bibinfo {author}
  {\bibfnamefont {S.~L.}\ \bibnamefont {Schreiber}}, \bibinfo {author}
  {\bibfnamefont {P.~A.}\ \bibnamefont {Clemons}}, \bibinfo {author}
  {\bibfnamefont {A.~E.}\ \bibnamefont {Carpenter}}, \ and\ \bibinfo {author}
  {\bibfnamefont {A.~F.}\ \bibnamefont {Shamji}},\ }\bibfield  {title}
  {{\selectlanguage{english}\enquote {\bibinfo {title} {Multiplex {Cytological}
  {Profiling} {Assay} to {Measure} {Diverse} {Cellular} {States}},}\ }}\href
  {\doibase 10.1371/journal.pone.0080999} {\bibfield  {journal} {\bibinfo
  {journal} {PLOS ONE}\ }\textbf {\bibinfo {volume} {8}},\ \bibinfo {pages}
  {e80999} (\bibinfo {year} {2013})},\ \bibinfo {note} {publisher: Public
  Library of Science}\BibitemShut {NoStop}%
\bibitem [{\citenamefont {Bray}\ \emph {et~al.}(2016)\citenamefont {Bray},
  \citenamefont {Singh}, \citenamefont {Han}, \citenamefont {Davis},
  \citenamefont {Borgeson}, \citenamefont {Hartland}, \citenamefont
  {Kost-Alimova}, \citenamefont {Gustafsdottir}, \citenamefont {Gibson},\ and\
  \citenamefont {Carpenter}}]{bray_cell_2016}%
  \BibitemOpen
  \bibfield  {author} {\bibinfo {author} {\bibfnamefont {M.-A.}\ \bibnamefont
  {Bray}}, \bibinfo {author} {\bibfnamefont {S.}~\bibnamefont {Singh}},
  \bibinfo {author} {\bibfnamefont {H.}~\bibnamefont {Han}}, \bibinfo {author}
  {\bibfnamefont {C.~T.}\ \bibnamefont {Davis}}, \bibinfo {author}
  {\bibfnamefont {B.}~\bibnamefont {Borgeson}}, \bibinfo {author}
  {\bibfnamefont {C.}~\bibnamefont {Hartland}}, \bibinfo {author}
  {\bibfnamefont {M.}~\bibnamefont {Kost-Alimova}}, \bibinfo {author}
  {\bibfnamefont {S.~M.}\ \bibnamefont {Gustafsdottir}}, \bibinfo {author}
  {\bibfnamefont {C.~C.}\ \bibnamefont {Gibson}}, \ and\ \bibinfo {author}
  {\bibfnamefont {A.~E.}\ \bibnamefont {Carpenter}},\ }\bibfield  {title}
  {\enquote {\bibinfo {title} {Cell {Painting}, a high-content image-based
  assay for morphological profiling using multiplexed fluorescent dyes},}\
  }\href {\doibase 10.1038/nprot.2016.105} {\bibfield  {journal} {\bibinfo
  {journal} {Nature protocols}\ }\textbf {\bibinfo {volume} {11}},\ \bibinfo
  {pages} {1757--1774} (\bibinfo {year} {2016})}\BibitemShut {NoStop}%
\bibitem [{\citenamefont {Cimini}\ \emph {et~al.}(2023)\citenamefont {Cimini},
  \citenamefont {Chandrasekaran}, \citenamefont {Kost-Alimova}, \citenamefont
  {Miller}, \citenamefont {Goodale}, \citenamefont {Fritchman}, \citenamefont
  {Byrne}, \citenamefont {Garg}, \citenamefont {Jamali}, \citenamefont {Logan},
  \citenamefont {Concannon}, \citenamefont {Lardeau}, \citenamefont {Mouchet},
  \citenamefont {Singh}, \citenamefont {Shafqat~Abbasi}, \citenamefont
  {Aspesi}, \citenamefont {Boyd}, \citenamefont {Gilbert}, \citenamefont
  {Gnutt}, \citenamefont {Hariharan}, \citenamefont {Hernandez}, \citenamefont
  {Hormel}, \citenamefont {Juhani}, \citenamefont {Melanson}, \citenamefont
  {Mervin}, \citenamefont {Monteverde}, \citenamefont {Pilling}, \citenamefont
  {Skepner}, \citenamefont {Swalley}, \citenamefont {Vrcic}, \citenamefont
  {Weisbart}, \citenamefont {Williams}, \citenamefont {Yu}, \citenamefont
  {Zapiec},\ and\ \citenamefont {Carpenter}}]{cimini_optimizing_2023}%
  \BibitemOpen
  \bibfield  {author} {\bibinfo {author} {\bibfnamefont {B.~A.}\ \bibnamefont
  {Cimini}}, \bibinfo {author} {\bibfnamefont {S.~N.}\ \bibnamefont
  {Chandrasekaran}}, \bibinfo {author} {\bibfnamefont {M.}~\bibnamefont
  {Kost-Alimova}}, \bibinfo {author} {\bibfnamefont {L.}~\bibnamefont
  {Miller}}, \bibinfo {author} {\bibfnamefont {A.}~\bibnamefont {Goodale}},
  \bibinfo {author} {\bibfnamefont {B.}~\bibnamefont {Fritchman}}, \bibinfo
  {author} {\bibfnamefont {P.}~\bibnamefont {Byrne}}, \bibinfo {author}
  {\bibfnamefont {S.}~\bibnamefont {Garg}}, \bibinfo {author} {\bibfnamefont
  {N.}~\bibnamefont {Jamali}}, \bibinfo {author} {\bibfnamefont {D.~J.}\
  \bibnamefont {Logan}}, \bibinfo {author} {\bibfnamefont {J.~B.}\ \bibnamefont
  {Concannon}}, \bibinfo {author} {\bibfnamefont {C.-H.}\ \bibnamefont
  {Lardeau}}, \bibinfo {author} {\bibfnamefont {E.}~\bibnamefont {Mouchet}},
  \bibinfo {author} {\bibfnamefont {S.}~\bibnamefont {Singh}}, \bibinfo
  {author} {\bibfnamefont {H.}~\bibnamefont {Shafqat~Abbasi}}, \bibinfo
  {author} {\bibfnamefont {P.}~\bibnamefont {Aspesi}}, \bibinfo {author}
  {\bibfnamefont {J.~D.}\ \bibnamefont {Boyd}}, \bibinfo {author}
  {\bibfnamefont {T.}~\bibnamefont {Gilbert}}, \bibinfo {author} {\bibfnamefont
  {D.}~\bibnamefont {Gnutt}}, \bibinfo {author} {\bibfnamefont
  {S.}~\bibnamefont {Hariharan}}, \bibinfo {author} {\bibfnamefont
  {D.}~\bibnamefont {Hernandez}}, \bibinfo {author} {\bibfnamefont
  {G.}~\bibnamefont {Hormel}}, \bibinfo {author} {\bibfnamefont
  {K.}~\bibnamefont {Juhani}}, \bibinfo {author} {\bibfnamefont
  {M.}~\bibnamefont {Melanson}}, \bibinfo {author} {\bibfnamefont {L.~H.}\
  \bibnamefont {Mervin}}, \bibinfo {author} {\bibfnamefont {T.}~\bibnamefont
  {Monteverde}}, \bibinfo {author} {\bibfnamefont {J.~E.}\ \bibnamefont
  {Pilling}}, \bibinfo {author} {\bibfnamefont {A.}~\bibnamefont {Skepner}},
  \bibinfo {author} {\bibfnamefont {S.~E.}\ \bibnamefont {Swalley}}, \bibinfo
  {author} {\bibfnamefont {A.}~\bibnamefont {Vrcic}}, \bibinfo {author}
  {\bibfnamefont {E.}~\bibnamefont {Weisbart}}, \bibinfo {author}
  {\bibfnamefont {G.}~\bibnamefont {Williams}}, \bibinfo {author}
  {\bibfnamefont {S.}~\bibnamefont {Yu}}, \bibinfo {author} {\bibfnamefont
  {B.}~\bibnamefont {Zapiec}}, \ and\ \bibinfo {author} {\bibfnamefont {A.~E.}\
  \bibnamefont {Carpenter}},\ }\bibfield  {title} {{\selectlanguage{english}\enquote {\bibinfo {title} {Optimizing the {Cell} {Painting} assay for
  image-based profiling},}\ }}\href {\doibase 10.1038/s41596-023-00840-9}
  {\bibfield  {journal} {\bibinfo  {journal} {Nature Protocols}\ }\textbf
  {\bibinfo {volume} {18}},\ \bibinfo {pages} {1981--2013} (\bibinfo {year}
  {2023})},\ \bibinfo {note} {publisher: Nature Publishing Group}\BibitemShut
  {NoStop}%
\bibitem [{\citenamefont {Ljosa}, \citenamefont {Sokolnicki},\ and\
  \citenamefont {Carpenter}(2013)}]{ljosa_correction_2013}%
  \BibitemOpen
  \bibfield  {author} {\bibinfo {author} {\bibfnamefont {V.}~\bibnamefont
  {Ljosa}}, \bibinfo {author} {\bibfnamefont {K.~L.}\ \bibnamefont
  {Sokolnicki}}, \ and\ \bibinfo {author} {\bibfnamefont {A.~E.}\ \bibnamefont
  {Carpenter}},\ }\bibfield  {title} {{\selectlanguage{english}\enquote {\bibinfo
  {title} {Correction: {Corrigendum}: {Annotated} high-throughput microscopy
  image sets for validation},}\ }}\href {\doibase 10.1038/nmeth0513-445d}
  {\bibfield  {journal} {\bibinfo  {journal} {Nature Methods}\ }\textbf
  {\bibinfo {volume} {10}},\ \bibinfo {pages} {445--445} (\bibinfo {year}
  {2013})},\ \bibinfo {note} {publisher: Nature Publishing Group}\BibitemShut
  {NoStop}%
\bibitem [{\citenamefont {Bray}\ \emph {et~al.}(2017)\citenamefont {Bray},
  \citenamefont {Gustafsdottir}, \citenamefont {Rohban}, \citenamefont {Singh},
  \citenamefont {Ljosa}, \citenamefont {Sokolnicki}, \citenamefont {Bittker},
  \citenamefont {Bodycombe}, \citenamefont {Dančík}, \citenamefont {Hasaka},
  \citenamefont {Hon}, \citenamefont {Kemp}, \citenamefont {Li}, \citenamefont
  {Walpita}, \citenamefont {Wawer}, \citenamefont {Golub}, \citenamefont
  {Schreiber}, \citenamefont {Clemons}, \citenamefont {Shamji},\ and\
  \citenamefont {Carpenter}}]{bray_dataset_2017}%
  \BibitemOpen
  \bibfield  {author} {\bibinfo {author} {\bibfnamefont {M.-A.}\ \bibnamefont
  {Bray}}, \bibinfo {author} {\bibfnamefont {S.~M.}\ \bibnamefont
  {Gustafsdottir}}, \bibinfo {author} {\bibfnamefont {M.~H.}\ \bibnamefont
  {Rohban}}, \bibinfo {author} {\bibfnamefont {S.}~\bibnamefont {Singh}},
  \bibinfo {author} {\bibfnamefont {V.}~\bibnamefont {Ljosa}}, \bibinfo
  {author} {\bibfnamefont {K.~L.}\ \bibnamefont {Sokolnicki}}, \bibinfo
  {author} {\bibfnamefont {J.~A.}\ \bibnamefont {Bittker}}, \bibinfo {author}
  {\bibfnamefont {N.~E.}\ \bibnamefont {Bodycombe}}, \bibinfo {author}
  {\bibfnamefont {V.}~\bibnamefont {Dančík}}, \bibinfo {author}
  {\bibfnamefont {T.~P.}\ \bibnamefont {Hasaka}}, \bibinfo {author}
  {\bibfnamefont {C.~S.}\ \bibnamefont {Hon}}, \bibinfo {author} {\bibfnamefont
  {M.~M.}\ \bibnamefont {Kemp}}, \bibinfo {author} {\bibfnamefont
  {K.}~\bibnamefont {Li}}, \bibinfo {author} {\bibfnamefont {D.}~\bibnamefont
  {Walpita}}, \bibinfo {author} {\bibfnamefont {M.~J.}\ \bibnamefont {Wawer}},
  \bibinfo {author} {\bibfnamefont {T.~R.}\ \bibnamefont {Golub}}, \bibinfo
  {author} {\bibfnamefont {S.~L.}\ \bibnamefont {Schreiber}}, \bibinfo {author}
  {\bibfnamefont {P.~A.}\ \bibnamefont {Clemons}}, \bibinfo {author}
  {\bibfnamefont {A.~F.}\ \bibnamefont {Shamji}}, \ and\ \bibinfo {author}
  {\bibfnamefont {A.~E.}\ \bibnamefont {Carpenter}},\ }\bibfield  {title}
  {\enquote {\bibinfo {title} {A dataset of images and morphological profiles
  of 30 000 small-molecule treatments using the {Cell} {Painting} assay},}\
  }\href {\doibase 10.1093/gigascience/giw014} {\bibfield  {journal} {\bibinfo
  {journal} {GigaScience}\ }\textbf {\bibinfo {volume} {6}},\ \bibinfo {pages}
  {giw014} (\bibinfo {year} {2017})}\BibitemShut {NoStop}%
\bibitem [{\citenamefont {Caicedo}\ \emph {et~al.}(2022)\citenamefont
  {Caicedo}, \citenamefont {Arevalo}, \citenamefont {Piccioni}, \citenamefont
  {Bray}, \citenamefont {Hartland}, \citenamefont {Wu}, \citenamefont {Brooks},
  \citenamefont {Berger}, \citenamefont {Boehm}, \citenamefont {Carpenter},\
  and\ \citenamefont {Singh}}]{caicedo_cell_2022}%
  \BibitemOpen
  \bibfield  {author} {\bibinfo {author} {\bibfnamefont {J.~C.}\ \bibnamefont
  {Caicedo}}, \bibinfo {author} {\bibfnamefont {J.}~\bibnamefont {Arevalo}},
  \bibinfo {author} {\bibfnamefont {F.}~\bibnamefont {Piccioni}}, \bibinfo
  {author} {\bibfnamefont {M.-A.}\ \bibnamefont {Bray}}, \bibinfo {author}
  {\bibfnamefont {C.~L.}\ \bibnamefont {Hartland}}, \bibinfo {author}
  {\bibfnamefont {X.}~\bibnamefont {Wu}}, \bibinfo {author} {\bibfnamefont
  {A.~N.}\ \bibnamefont {Brooks}}, \bibinfo {author} {\bibfnamefont {A.~H.}\
  \bibnamefont {Berger}}, \bibinfo {author} {\bibfnamefont {J.~S.}\
  \bibnamefont {Boehm}}, \bibinfo {author} {\bibfnamefont {A.~E.}\ \bibnamefont
  {Carpenter}}, \ and\ \bibinfo {author} {\bibfnamefont {S.}~\bibnamefont
  {Singh}},\ }\bibfield  {title} {\enquote {\bibinfo {title} {Cell {Painting}
  predicts impact of lung cancer variants},}\ }\href {\doibase
  10.1091/mbc.E21-11-0538} {\bibfield  {journal} {\bibinfo  {journal}
  {Molecular Biology of the Cell}\ }\textbf {\bibinfo {volume} {33}},\ \bibinfo
  {pages} {ar49} (\bibinfo {year} {2022})},\ \bibinfo {note} {publisher:
  American Society for Cell Biology (mboc)}\BibitemShut {NoStop}%
\bibitem [{\citenamefont {Chandrasekaran}\ \emph {et~al.}(2023)\citenamefont
  {Chandrasekaran}, \citenamefont {Ackerman}, \citenamefont {Alix},
  \citenamefont {Ando}, \citenamefont {Arevalo}, \citenamefont {Bennion},
  \citenamefont {Boisseau}, \citenamefont {Borowa}, \citenamefont {Boyd},
  \citenamefont {Brino}, \citenamefont {Byrne}, \citenamefont {Ceulemans},
  \citenamefont {Ch’ng}, \citenamefont {Cimini}, \citenamefont {Clevert},
  \citenamefont {Deflaux}, \citenamefont {Doench}, \citenamefont {Dorval},
  \citenamefont {Doyonnas}, \citenamefont {Dragone}, \citenamefont {Engkvist},
  \citenamefont {Faloon}, \citenamefont {Fritchman}, \citenamefont {Fuchs},
  \citenamefont {Garg}, \citenamefont {Gilbert}, \citenamefont {Glazer},
  \citenamefont {Gnutt}, \citenamefont {Goodale}, \citenamefont {Grignard},
  \citenamefont {Guenther}, \citenamefont {Han}, \citenamefont {Hanifehlou},
  \citenamefont {Hariharan}, \citenamefont {Hernandez}, \citenamefont {Horman},
  \citenamefont {Hormel}, \citenamefont {Huntley}, \citenamefont {Icke},
  \citenamefont {Iida}, \citenamefont {Jacob}, \citenamefont {Jaensch},
  \citenamefont {Khetan}, \citenamefont {Kost-Alimova}, \citenamefont
  {Krawiec}, \citenamefont {Kuhn}, \citenamefont {Lardeau}, \citenamefont
  {Lembke}, \citenamefont {Lin}, \citenamefont {Little}, \citenamefont
  {Lofstrom}, \citenamefont {Lotfi}, \citenamefont {Logan}, \citenamefont
  {Luo}, \citenamefont {Madoux}, \citenamefont {Zapata}, \citenamefont
  {Marion}, \citenamefont {Martin}, \citenamefont {McCarthy}, \citenamefont
  {Mervin}, \citenamefont {Miller}, \citenamefont {Mohamed}, \citenamefont
  {Monteverde}, \citenamefont {Mouchet}, \citenamefont {Nicke}, \citenamefont
  {Ogier}, \citenamefont {Ong}, \citenamefont {Osterland}, \citenamefont
  {Otrocka}, \citenamefont {Peeters}, \citenamefont {Pilling}, \citenamefont
  {Prechtl}, \citenamefont {Qian}, \citenamefont {Rataj}, \citenamefont {Root},
  \citenamefont {Sakata}, \citenamefont {Scrace}, \citenamefont {Shimizu},
  \citenamefont {Simon}, \citenamefont {Sommer}, \citenamefont {Spruiell},
  \citenamefont {Sumia}, \citenamefont {Swalley}, \citenamefont {Terauchi},
  \citenamefont {Thibaudeau}, \citenamefont {Unruh}, \citenamefont {Waeter},
  \citenamefont {Dyck}, \citenamefont {Staden}, \citenamefont {Warchoł},
  \citenamefont {Weisbart}, \citenamefont {Weiss}, \citenamefont
  {Wiest-Daessle}, \citenamefont {Williams}, \citenamefont {Yu}, \citenamefont
  {Zapiec}, \citenamefont {Żyła}, \citenamefont {Singh},\ and\ \citenamefont
  {Carpenter}}]{chandrasekaran_jump_2023}%
  \BibitemOpen
  \bibfield  {author} {\bibinfo {author} {\bibfnamefont {S.~N.}\ \bibnamefont
  {Chandrasekaran}}, \bibinfo {author} {\bibfnamefont {J.}~\bibnamefont
  {Ackerman}}, \bibinfo {author} {\bibfnamefont {E.}~\bibnamefont {Alix}},
  \bibinfo {author} {\bibfnamefont {D.~M.}\ \bibnamefont {Ando}}, \bibinfo
  {author} {\bibfnamefont {J.}~\bibnamefont {Arevalo}}, \bibinfo {author}
  {\bibfnamefont {M.}~\bibnamefont {Bennion}}, \bibinfo {author} {\bibfnamefont
  {N.}~\bibnamefont {Boisseau}}, \bibinfo {author} {\bibfnamefont
  {A.}~\bibnamefont {Borowa}}, \bibinfo {author} {\bibfnamefont {J.~D.}\
  \bibnamefont {Boyd}}, \bibinfo {author} {\bibfnamefont {L.}~\bibnamefont
  {Brino}}, \bibinfo {author} {\bibfnamefont {P.~J.}\ \bibnamefont {Byrne}},
  \bibinfo {author} {\bibfnamefont {H.}~\bibnamefont {Ceulemans}}, \bibinfo
  {author} {\bibfnamefont {C.}~\bibnamefont {Ch’ng}}, \bibinfo {author}
  {\bibfnamefont {B.~A.}\ \bibnamefont {Cimini}}, \bibinfo {author}
  {\bibfnamefont {D.-A.}\ \bibnamefont {Clevert}}, \bibinfo {author}
  {\bibfnamefont {N.}~\bibnamefont {Deflaux}}, \bibinfo {author} {\bibfnamefont
  {J.~G.}\ \bibnamefont {Doench}}, \bibinfo {author} {\bibfnamefont
  {T.}~\bibnamefont {Dorval}}, \bibinfo {author} {\bibfnamefont
  {R.}~\bibnamefont {Doyonnas}}, \bibinfo {author} {\bibfnamefont
  {V.}~\bibnamefont {Dragone}}, \bibinfo {author} {\bibfnamefont
  {O.}~\bibnamefont {Engkvist}}, \bibinfo {author} {\bibfnamefont {P.~W.}\
  \bibnamefont {Faloon}}, \bibinfo {author} {\bibfnamefont {B.}~\bibnamefont
  {Fritchman}}, \bibinfo {author} {\bibfnamefont {F.}~\bibnamefont {Fuchs}},
  \bibinfo {author} {\bibfnamefont {S.}~\bibnamefont {Garg}}, \bibinfo {author}
  {\bibfnamefont {T.~J.}\ \bibnamefont {Gilbert}}, \bibinfo {author}
  {\bibfnamefont {D.}~\bibnamefont {Glazer}}, \bibinfo {author} {\bibfnamefont
  {D.}~\bibnamefont {Gnutt}}, \bibinfo {author} {\bibfnamefont
  {A.}~\bibnamefont {Goodale}}, \bibinfo {author} {\bibfnamefont
  {J.}~\bibnamefont {Grignard}}, \bibinfo {author} {\bibfnamefont
  {J.}~\bibnamefont {Guenther}}, \bibinfo {author} {\bibfnamefont
  {Y.}~\bibnamefont {Han}}, \bibinfo {author} {\bibfnamefont {Z.}~\bibnamefont
  {Hanifehlou}}, \bibinfo {author} {\bibfnamefont {S.}~\bibnamefont
  {Hariharan}}, \bibinfo {author} {\bibfnamefont {D.}~\bibnamefont
  {Hernandez}}, \bibinfo {author} {\bibfnamefont {S.~R.}\ \bibnamefont
  {Horman}}, \bibinfo {author} {\bibfnamefont {G.}~\bibnamefont {Hormel}},
  \bibinfo {author} {\bibfnamefont {M.}~\bibnamefont {Huntley}}, \bibinfo
  {author} {\bibfnamefont {I.}~\bibnamefont {Icke}}, \bibinfo {author}
  {\bibfnamefont {M.}~\bibnamefont {Iida}}, \bibinfo {author} {\bibfnamefont
  {C.~B.}\ \bibnamefont {Jacob}}, \bibinfo {author} {\bibfnamefont
  {S.}~\bibnamefont {Jaensch}}, \bibinfo {author} {\bibfnamefont
  {J.}~\bibnamefont {Khetan}}, \bibinfo {author} {\bibfnamefont
  {M.}~\bibnamefont {Kost-Alimova}}, \bibinfo {author} {\bibfnamefont
  {T.}~\bibnamefont {Krawiec}}, \bibinfo {author} {\bibfnamefont
  {D.}~\bibnamefont {Kuhn}}, \bibinfo {author} {\bibfnamefont {C.-H.}\
  \bibnamefont {Lardeau}}, \bibinfo {author} {\bibfnamefont {A.}~\bibnamefont
  {Lembke}}, \bibinfo {author} {\bibfnamefont {F.}~\bibnamefont {Lin}},
  \bibinfo {author} {\bibfnamefont {K.~D.}\ \bibnamefont {Little}}, \bibinfo
  {author} {\bibfnamefont {K.~R.}\ \bibnamefont {Lofstrom}}, \bibinfo {author}
  {\bibfnamefont {S.}~\bibnamefont {Lotfi}}, \bibinfo {author} {\bibfnamefont
  {D.~J.}\ \bibnamefont {Logan}}, \bibinfo {author} {\bibfnamefont
  {Y.}~\bibnamefont {Luo}}, \bibinfo {author} {\bibfnamefont {F.}~\bibnamefont
  {Madoux}}, \bibinfo {author} {\bibfnamefont {P.~A.~M.}\ \bibnamefont
  {Zapata}}, \bibinfo {author} {\bibfnamefont {B.~A.}\ \bibnamefont {Marion}},
  \bibinfo {author} {\bibfnamefont {G.}~\bibnamefont {Martin}}, \bibinfo
  {author} {\bibfnamefont {N.~J.}\ \bibnamefont {McCarthy}}, \bibinfo {author}
  {\bibfnamefont {L.}~\bibnamefont {Mervin}}, \bibinfo {author} {\bibfnamefont
  {L.}~\bibnamefont {Miller}}, \bibinfo {author} {\bibfnamefont
  {H.}~\bibnamefont {Mohamed}}, \bibinfo {author} {\bibfnamefont
  {T.}~\bibnamefont {Monteverde}}, \bibinfo {author} {\bibfnamefont
  {E.}~\bibnamefont {Mouchet}}, \bibinfo {author} {\bibfnamefont
  {B.}~\bibnamefont {Nicke}}, \bibinfo {author} {\bibfnamefont
  {A.}~\bibnamefont {Ogier}}, \bibinfo {author} {\bibfnamefont {A.-L.}\
  \bibnamefont {Ong}}, \bibinfo {author} {\bibfnamefont {M.}~\bibnamefont
  {Osterland}}, \bibinfo {author} {\bibfnamefont {M.}~\bibnamefont {Otrocka}},
  \bibinfo {author} {\bibfnamefont {P.~J.}\ \bibnamefont {Peeters}}, \bibinfo
  {author} {\bibfnamefont {J.}~\bibnamefont {Pilling}}, \bibinfo {author}
  {\bibfnamefont {S.}~\bibnamefont {Prechtl}}, \bibinfo {author} {\bibfnamefont
  {C.}~\bibnamefont {Qian}}, \bibinfo {author} {\bibfnamefont {K.}~\bibnamefont
  {Rataj}}, \bibinfo {author} {\bibfnamefont {D.~E.}\ \bibnamefont {Root}},
  \bibinfo {author} {\bibfnamefont {S.~K.}\ \bibnamefont {Sakata}}, \bibinfo
  {author} {\bibfnamefont {S.}~\bibnamefont {Scrace}}, \bibinfo {author}
  {\bibfnamefont {H.}~\bibnamefont {Shimizu}}, \bibinfo {author} {\bibfnamefont
  {D.}~\bibnamefont {Simon}}, \bibinfo {author} {\bibfnamefont
  {P.}~\bibnamefont {Sommer}}, \bibinfo {author} {\bibfnamefont
  {C.}~\bibnamefont {Spruiell}}, \bibinfo {author} {\bibfnamefont
  {I.}~\bibnamefont {Sumia}}, \bibinfo {author} {\bibfnamefont {S.~E.}\
  \bibnamefont {Swalley}}, \bibinfo {author} {\bibfnamefont {H.}~\bibnamefont
  {Terauchi}}, \bibinfo {author} {\bibfnamefont {A.}~\bibnamefont
  {Thibaudeau}}, \bibinfo {author} {\bibfnamefont {A.}~\bibnamefont {Unruh}},
  \bibinfo {author} {\bibfnamefont {J.~V.~d.}\ \bibnamefont {Waeter}}, \bibinfo
  {author} {\bibfnamefont {M.~V.}\ \bibnamefont {Dyck}}, \bibinfo {author}
  {\bibfnamefont {C.~v.}\ \bibnamefont {Staden}}, \bibinfo {author}
  {\bibfnamefont {M.}~\bibnamefont {Warchoł}}, \bibinfo {author}
  {\bibfnamefont {E.}~\bibnamefont {Weisbart}}, \bibinfo {author}
  {\bibfnamefont {A.}~\bibnamefont {Weiss}}, \bibinfo {author} {\bibfnamefont
  {N.}~\bibnamefont {Wiest-Daessle}}, \bibinfo {author} {\bibfnamefont
  {G.}~\bibnamefont {Williams}}, \bibinfo {author} {\bibfnamefont
  {S.}~\bibnamefont {Yu}}, \bibinfo {author} {\bibfnamefont {B.}~\bibnamefont
  {Zapiec}}, \bibinfo {author} {\bibfnamefont {M.}~\bibnamefont {Żyła}},
  \bibinfo {author} {\bibfnamefont {S.}~\bibnamefont {Singh}}, \ and\ \bibinfo
  {author} {\bibfnamefont {A.~E.}\ \bibnamefont {Carpenter}},\ }\href {\doibase
  10.1101/2023.03.23.534023} {{\selectlanguage{english}\enquote {\bibinfo {title}
  {{JUMP} {Cell} {Painting} dataset: morphological impact of 136,000 chemical
  and genetic perturbations},}\ }} (\bibinfo {year} {2023}),\ \bibinfo {note}
  {pages: 2023.03.23.534023 Section: New Results}\BibitemShut {NoStop}%
\bibitem [{\citenamefont {Thul}\ \emph {et~al.}(2017)\citenamefont {Thul},
  \citenamefont {Åkesson}, \citenamefont {Wiking}, \citenamefont {Mahdessian},
  \citenamefont {Geladaki}, \citenamefont {Ait~Blal}, \citenamefont {Alm},
  \citenamefont {Asplund}, \citenamefont {Björk}, \citenamefont {Breckels},
  \citenamefont {Bäckström}, \citenamefont {Danielsson}, \citenamefont
  {Fagerberg}, \citenamefont {Fall}, \citenamefont {Gatto}, \citenamefont
  {Gnann}, \citenamefont {Hober}, \citenamefont {Hjelmare}, \citenamefont
  {Johansson}, \citenamefont {Lee}, \citenamefont {Lindskog}, \citenamefont
  {Mulder}, \citenamefont {Mulvey}, \citenamefont {Nilsson}, \citenamefont
  {Oksvold}, \citenamefont {Rockberg}, \citenamefont {Schutten}, \citenamefont
  {Schwenk}, \citenamefont {Sivertsson}, \citenamefont {Sjöstedt},
  \citenamefont {Skogs}, \citenamefont {Stadler}, \citenamefont {Sullivan},
  \citenamefont {Tegel}, \citenamefont {Winsnes}, \citenamefont {Zhang},
  \citenamefont {Zwahlen}, \citenamefont {Mardinoglu}, \citenamefont {Pontén},
  \citenamefont {von Feilitzen}, \citenamefont {Lilley}, \citenamefont
  {Uhlén},\ and\ \citenamefont {Lundberg}}]{thul_subcellular_2017}%
  \BibitemOpen
  \bibfield  {author} {\bibinfo {author} {\bibfnamefont {P.~J.}\ \bibnamefont
  {Thul}}, \bibinfo {author} {\bibfnamefont {L.}~\bibnamefont {Åkesson}},
  \bibinfo {author} {\bibfnamefont {M.}~\bibnamefont {Wiking}}, \bibinfo
  {author} {\bibfnamefont {D.}~\bibnamefont {Mahdessian}}, \bibinfo {author}
  {\bibfnamefont {A.}~\bibnamefont {Geladaki}}, \bibinfo {author}
  {\bibfnamefont {H.}~\bibnamefont {Ait~Blal}}, \bibinfo {author}
  {\bibfnamefont {T.}~\bibnamefont {Alm}}, \bibinfo {author} {\bibfnamefont
  {A.}~\bibnamefont {Asplund}}, \bibinfo {author} {\bibfnamefont
  {L.}~\bibnamefont {Björk}}, \bibinfo {author} {\bibfnamefont {L.~M.}\
  \bibnamefont {Breckels}}, \bibinfo {author} {\bibfnamefont {A.}~\bibnamefont
  {Bäckström}}, \bibinfo {author} {\bibfnamefont {F.}~\bibnamefont
  {Danielsson}}, \bibinfo {author} {\bibfnamefont {L.}~\bibnamefont
  {Fagerberg}}, \bibinfo {author} {\bibfnamefont {J.}~\bibnamefont {Fall}},
  \bibinfo {author} {\bibfnamefont {L.}~\bibnamefont {Gatto}}, \bibinfo
  {author} {\bibfnamefont {C.}~\bibnamefont {Gnann}}, \bibinfo {author}
  {\bibfnamefont {S.}~\bibnamefont {Hober}}, \bibinfo {author} {\bibfnamefont
  {M.}~\bibnamefont {Hjelmare}}, \bibinfo {author} {\bibfnamefont
  {F.}~\bibnamefont {Johansson}}, \bibinfo {author} {\bibfnamefont
  {S.}~\bibnamefont {Lee}}, \bibinfo {author} {\bibfnamefont {C.}~\bibnamefont
  {Lindskog}}, \bibinfo {author} {\bibfnamefont {J.}~\bibnamefont {Mulder}},
  \bibinfo {author} {\bibfnamefont {C.~M.}\ \bibnamefont {Mulvey}}, \bibinfo
  {author} {\bibfnamefont {P.}~\bibnamefont {Nilsson}}, \bibinfo {author}
  {\bibfnamefont {P.}~\bibnamefont {Oksvold}}, \bibinfo {author} {\bibfnamefont
  {J.}~\bibnamefont {Rockberg}}, \bibinfo {author} {\bibfnamefont
  {R.}~\bibnamefont {Schutten}}, \bibinfo {author} {\bibfnamefont {J.~M.}\
  \bibnamefont {Schwenk}}, \bibinfo {author} {\bibfnamefont
  {{\AA}.}~\bibnamefont {Sivertsson}}, \bibinfo {author} {\bibfnamefont
  {E.}~\bibnamefont {Sjöstedt}}, \bibinfo {author} {\bibfnamefont
  {M.}~\bibnamefont {Skogs}}, \bibinfo {author} {\bibfnamefont
  {C.}~\bibnamefont {Stadler}}, \bibinfo {author} {\bibfnamefont {D.~P.}\
  \bibnamefont {Sullivan}}, \bibinfo {author} {\bibfnamefont {H.}~\bibnamefont
  {Tegel}}, \bibinfo {author} {\bibfnamefont {C.}~\bibnamefont {Winsnes}},
  \bibinfo {author} {\bibfnamefont {C.}~\bibnamefont {Zhang}}, \bibinfo
  {author} {\bibfnamefont {M.}~\bibnamefont {Zwahlen}}, \bibinfo {author}
  {\bibfnamefont {A.}~\bibnamefont {Mardinoglu}}, \bibinfo {author}
  {\bibfnamefont {F.}~\bibnamefont {Pontén}}, \bibinfo {author} {\bibfnamefont
  {K.}~\bibnamefont {von Feilitzen}}, \bibinfo {author} {\bibfnamefont {K.~S.}\
  \bibnamefont {Lilley}}, \bibinfo {author} {\bibfnamefont {M.}~\bibnamefont
  {Uhlén}}, \ and\ \bibinfo {author} {\bibfnamefont {E.}~\bibnamefont
  {Lundberg}},\ }\bibfield  {title} {\enquote {\bibinfo {title} {A subcellular
  map of the human proteome},}\ }\href {\doibase 10.1126/science.aal3321}
  {\bibfield  {journal} {\bibinfo  {journal} {Science}\ }\textbf {\bibinfo
  {volume} {356}},\ \bibinfo {pages} {eaal3321} (\bibinfo {year} {2017})},\
  \bibinfo {note} {publisher: American Association for the Advancement of
  Science}\BibitemShut {NoStop}%
\bibitem [{\citenamefont {Cho}\ \emph {et~al.}(2022)\citenamefont {Cho},
  \citenamefont {Cheveralls}, \citenamefont {Brunner}, \citenamefont {Kim},
  \citenamefont {Michaelis}, \citenamefont {Raghavan}, \citenamefont
  {Kobayashi}, \citenamefont {Savy}, \citenamefont {Li}, \citenamefont {Canaj},
  \citenamefont {Kim}, \citenamefont {Stewart}, \citenamefont {Gnann},
  \citenamefont {McCarthy}, \citenamefont {Cabrera}, \citenamefont {Brunetti},
  \citenamefont {Chhun}, \citenamefont {Dingle}, \citenamefont {Hein},
  \citenamefont {Huang}, \citenamefont {Mehta}, \citenamefont {Weissman},
  \citenamefont {Gómez-Sjöberg}, \citenamefont {Itzhak}, \citenamefont
  {Royer}, \citenamefont {Mann},\ and\ \citenamefont
  {Leonetti}}]{cho_opencell_2022}%
  \BibitemOpen
  \bibfield  {author} {\bibinfo {author} {\bibfnamefont {N.~H.}\ \bibnamefont
  {Cho}}, \bibinfo {author} {\bibfnamefont {K.~C.}\ \bibnamefont {Cheveralls}},
  \bibinfo {author} {\bibfnamefont {A.-D.}\ \bibnamefont {Brunner}}, \bibinfo
  {author} {\bibfnamefont {K.}~\bibnamefont {Kim}}, \bibinfo {author}
  {\bibfnamefont {A.~C.}\ \bibnamefont {Michaelis}}, \bibinfo {author}
  {\bibfnamefont {P.}~\bibnamefont {Raghavan}}, \bibinfo {author}
  {\bibfnamefont {H.}~\bibnamefont {Kobayashi}}, \bibinfo {author}
  {\bibfnamefont {L.}~\bibnamefont {Savy}}, \bibinfo {author} {\bibfnamefont
  {J.~Y.}\ \bibnamefont {Li}}, \bibinfo {author} {\bibfnamefont
  {H.}~\bibnamefont {Canaj}}, \bibinfo {author} {\bibfnamefont {J.~Y.~S.}\
  \bibnamefont {Kim}}, \bibinfo {author} {\bibfnamefont {E.~M.}\ \bibnamefont
  {Stewart}}, \bibinfo {author} {\bibfnamefont {C.}~\bibnamefont {Gnann}},
  \bibinfo {author} {\bibfnamefont {F.}~\bibnamefont {McCarthy}}, \bibinfo
  {author} {\bibfnamefont {J.~P.}\ \bibnamefont {Cabrera}}, \bibinfo {author}
  {\bibfnamefont {R.~M.}\ \bibnamefont {Brunetti}}, \bibinfo {author}
  {\bibfnamefont {B.~B.}\ \bibnamefont {Chhun}}, \bibinfo {author}
  {\bibfnamefont {G.}~\bibnamefont {Dingle}}, \bibinfo {author} {\bibfnamefont
  {M.~Y.}\ \bibnamefont {Hein}}, \bibinfo {author} {\bibfnamefont
  {B.}~\bibnamefont {Huang}}, \bibinfo {author} {\bibfnamefont {S.~B.}\
  \bibnamefont {Mehta}}, \bibinfo {author} {\bibfnamefont {J.~S.}\ \bibnamefont
  {Weissman}}, \bibinfo {author} {\bibfnamefont {R.}~\bibnamefont
  {Gómez-Sjöberg}}, \bibinfo {author} {\bibfnamefont {D.~N.}\ \bibnamefont
  {Itzhak}}, \bibinfo {author} {\bibfnamefont {L.~A.}\ \bibnamefont {Royer}},
  \bibinfo {author} {\bibfnamefont {M.}~\bibnamefont {Mann}}, \ and\ \bibinfo
  {author} {\bibfnamefont {M.~D.}\ \bibnamefont {Leonetti}},\ }\bibfield
  {title} {\enquote {\bibinfo {title} {{OpenCell}: {Endogenous} tagging for the
  cartography of human cellular organization},}\ }\href {\doibase
  10.1126/science.abi6983} {\bibfield  {journal} {\bibinfo  {journal}
  {Science}\ }\textbf {\bibinfo {volume} {375}},\ \bibinfo {pages} {eabi6983}
  (\bibinfo {year} {2022})},\ \bibinfo {note} {publisher: American Association
  for the Advancement of Science}\BibitemShut {NoStop}%
\bibitem [{\citenamefont {Carpenter}\ \emph {et~al.}(2006)\citenamefont
  {Carpenter}, \citenamefont {Jones}, \citenamefont {Lamprecht}, \citenamefont
  {Clarke}, \citenamefont {Kang}, \citenamefont {Friman}, \citenamefont
  {Guertin}, \citenamefont {Chang}, \citenamefont {Lindquist}, \citenamefont
  {Moffat}, \citenamefont {Golland},\ and\ \citenamefont
  {Sabatini}}]{carpenter_cellprofiler_2006}%
  \BibitemOpen
  \bibfield  {author} {\bibinfo {author} {\bibfnamefont {A.~E.}\ \bibnamefont
  {Carpenter}}, \bibinfo {author} {\bibfnamefont {T.~R.}\ \bibnamefont
  {Jones}}, \bibinfo {author} {\bibfnamefont {M.~R.}\ \bibnamefont
  {Lamprecht}}, \bibinfo {author} {\bibfnamefont {C.}~\bibnamefont {Clarke}},
  \bibinfo {author} {\bibfnamefont {I.~H.}\ \bibnamefont {Kang}}, \bibinfo
  {author} {\bibfnamefont {O.}~\bibnamefont {Friman}}, \bibinfo {author}
  {\bibfnamefont {D.~A.}\ \bibnamefont {Guertin}}, \bibinfo {author}
  {\bibfnamefont {J.~H.}\ \bibnamefont {Chang}}, \bibinfo {author}
  {\bibfnamefont {R.~A.}\ \bibnamefont {Lindquist}}, \bibinfo {author}
  {\bibfnamefont {J.}~\bibnamefont {Moffat}}, \bibinfo {author} {\bibfnamefont
  {P.}~\bibnamefont {Golland}}, \ and\ \bibinfo {author} {\bibfnamefont
  {D.~M.}\ \bibnamefont {Sabatini}},\ }\bibfield  {title} {\enquote {\bibinfo
  {title} {{CellProfiler}: image analysis software for identifying and
  quantifying cell phenotypes},}\ }\href {\doibase 10.1186/gb-2006-7-10-r100}
  {\bibfield  {journal} {\bibinfo  {journal} {Genome Biology}\ }\textbf
  {\bibinfo {volume} {7}},\ \bibinfo {pages} {R100} (\bibinfo {year}
  {2006})}\BibitemShut {NoStop}%
\bibitem [{\citenamefont {Caicedo}, \citenamefont {Singh},\ and\ \citenamefont
  {Carpenter}(2016)}]{caicedo_applications_2016}%
  \BibitemOpen
  \bibfield  {author} {\bibinfo {author} {\bibfnamefont {J.~C.}\ \bibnamefont
  {Caicedo}}, \bibinfo {author} {\bibfnamefont {S.}~\bibnamefont {Singh}}, \
  and\ \bibinfo {author} {\bibfnamefont {A.~E.}\ \bibnamefont {Carpenter}},\
  }\bibfield  {title} {\enquote {\bibinfo {title} {Applications in image-based
  profiling of perturbations},}\ }\href {\doibase 10.1016/j.copbio.2016.04.003}
  {\bibfield  {journal} {\bibinfo  {journal} {Current Opinion in
  Biotechnology}\ }\bibinfo {series} {Systems biology •
  {Nanobiotechnology}},\ \textbf {\bibinfo {volume} {39}},\ \bibinfo {pages}
  {134--142} (\bibinfo {year} {2016})}\BibitemShut {NoStop}%
\bibitem [{\citenamefont {Pratapa}, \citenamefont {Doron},\ and\ \citenamefont
  {Caicedo}(2021)}]{pratapa_image-based_2021}%
  \BibitemOpen
  \bibfield  {author} {\bibinfo {author} {\bibfnamefont {A.}~\bibnamefont
  {Pratapa}}, \bibinfo {author} {\bibfnamefont {M.}~\bibnamefont {Doron}}, \
  and\ \bibinfo {author} {\bibfnamefont {J.~C.}\ \bibnamefont {Caicedo}},\
  }\bibfield  {title} {\enquote {\bibinfo {title} {Image-based cell phenotyping
  with deep learning},}\ }\href {\doibase 10.1016/j.cbpa.2021.04.001}
  {\bibfield  {journal} {\bibinfo  {journal} {Current Opinion in Chemical
  Biology}\ }\bibinfo {series} {Mechanistic {Biology} * {Machine} {Learning} in
  {Chemical} {Biology}},\ \textbf {\bibinfo {volume} {65}},\ \bibinfo {pages}
  {9--17} (\bibinfo {year} {2021})}\BibitemShut {NoStop}%
\bibitem [{\citenamefont {Tang}\ \emph {et~al.}(2024)\citenamefont {Tang},
  \citenamefont {Ratnayake}, \citenamefont {Seabra}, \citenamefont {Jiang},
  \citenamefont {Fang}, \citenamefont {Cui}, \citenamefont {Ding},
  \citenamefont {Kahveci}, \citenamefont {Bian}, \citenamefont {Li},
  \citenamefont {Luesch},\ and\ \citenamefont {Li}}]{tang_morphological_2024}%
  \BibitemOpen
  \bibfield  {author} {\bibinfo {author} {\bibfnamefont {Q.}~\bibnamefont
  {Tang}}, \bibinfo {author} {\bibfnamefont {R.}~\bibnamefont {Ratnayake}},
  \bibinfo {author} {\bibfnamefont {G.}~\bibnamefont {Seabra}}, \bibinfo
  {author} {\bibfnamefont {Z.}~\bibnamefont {Jiang}}, \bibinfo {author}
  {\bibfnamefont {R.}~\bibnamefont {Fang}}, \bibinfo {author} {\bibfnamefont
  {L.}~\bibnamefont {Cui}}, \bibinfo {author} {\bibfnamefont {Y.}~\bibnamefont
  {Ding}}, \bibinfo {author} {\bibfnamefont {T.}~\bibnamefont {Kahveci}},
  \bibinfo {author} {\bibfnamefont {J.}~\bibnamefont {Bian}}, \bibinfo {author}
  {\bibfnamefont {C.}~\bibnamefont {Li}}, \bibinfo {author} {\bibfnamefont
  {H.}~\bibnamefont {Luesch}}, \ and\ \bibinfo {author} {\bibfnamefont
  {Y.}~\bibnamefont {Li}},\ }\bibfield  {title} {\enquote {\bibinfo {title}
  {Morphological profiling for drug discovery in the era of deep learning},}\
  }\href {\doibase 10.1093/bib/bbae284} {\bibfield  {journal} {\bibinfo
  {journal} {Briefings in Bioinformatics}\ }\textbf {\bibinfo {volume} {25}},\
  \bibinfo {pages} {bbae284} (\bibinfo {year} {2024})}\BibitemShut {NoStop}%
\bibitem [{\citenamefont {Ronneberger}, \citenamefont {Fischer},\ and\
  \citenamefont {Brox}(2015)}]{ronneberger_u-net_2015}%
  \BibitemOpen
  \bibfield  {author} {\bibinfo {author} {\bibfnamefont {O.}~\bibnamefont
  {Ronneberger}}, \bibinfo {author} {\bibfnamefont {P.}~\bibnamefont
  {Fischer}}, \ and\ \bibinfo {author} {\bibfnamefont {T.}~\bibnamefont
  {Brox}},\ }\bibfield  {title} {{\selectlanguage{english}\enquote {\bibinfo
  {title} {U-{Net}: {Convolutional} {Networks} for {Biomedical} {Image}
  {Segmentation}},}\ }}in\ \href {\doibase 10.1007/978-3-319-24574-4_28}
  {{\selectlanguage{english}\emph {\bibinfo {booktitle} {Medical {Image}
  {Computing} and {Computer}-{Assisted} {Intervention} – {MICCAI} 2015}}}},\
  \bibinfo {editor} {edited by\ \bibinfo {editor} {\bibfnamefont
  {N.}~\bibnamefont {Navab}}, \bibinfo {editor} {\bibfnamefont
  {J.}~\bibnamefont {Hornegger}}, \bibinfo {editor} {\bibfnamefont {W.~M.}\
  \bibnamefont {Wells}}, \ and\ \bibinfo {editor} {\bibfnamefont {A.~F.}\
  \bibnamefont {Frangi}}}\ (\bibinfo  {publisher} {Springer International
  Publishing},\ \bibinfo {address} {Cham},\ \bibinfo {year} {2015})\ pp.\
  \bibinfo {pages} {234--241}\BibitemShut {NoStop}%
\bibitem [{\citenamefont {Hörst}\ \emph {et~al.}(2023)\citenamefont {Hörst},
  \citenamefont {Rempe}, \citenamefont {Heine}, \citenamefont {Seibold},
  \citenamefont {Keyl}, \citenamefont {Baldini}, \citenamefont {Ugurel},
  \citenamefont {Siveke}, \citenamefont {Grünwald}, \citenamefont {Egger},\
  and\ \citenamefont {Kleesiek}}]{horst_cellvit_2023}%
  \BibitemOpen
  \bibfield  {author} {\bibinfo {author} {\bibfnamefont {F.}~\bibnamefont
  {Hörst}}, \bibinfo {author} {\bibfnamefont {M.}~\bibnamefont {Rempe}},
  \bibinfo {author} {\bibfnamefont {L.}~\bibnamefont {Heine}}, \bibinfo
  {author} {\bibfnamefont {C.}~\bibnamefont {Seibold}}, \bibinfo {author}
  {\bibfnamefont {J.}~\bibnamefont {Keyl}}, \bibinfo {author} {\bibfnamefont
  {G.}~\bibnamefont {Baldini}}, \bibinfo {author} {\bibfnamefont
  {S.}~\bibnamefont {Ugurel}}, \bibinfo {author} {\bibfnamefont
  {J.}~\bibnamefont {Siveke}}, \bibinfo {author} {\bibfnamefont
  {B.}~\bibnamefont {Grünwald}}, \bibinfo {author} {\bibfnamefont
  {J.}~\bibnamefont {Egger}}, \ and\ \bibinfo {author} {\bibfnamefont
  {J.}~\bibnamefont {Kleesiek}},\ }\href {\doibase 10.48550/arXiv.2306.15350}
  {\enquote {\bibinfo {title} {{CellViT}: {Vision} {Transformers} for {Precise}
  {Cell} {Segmentation} and {Classification}},}\ } (\bibinfo {year} {2023}),\
  \bibinfo {note} {arXiv:2306.15350 [cs, eess]}\BibitemShut {NoStop}%
\bibitem [{\citenamefont {Minaee}\ \emph {et~al.}(2022)\citenamefont {Minaee},
  \citenamefont {Boykov}, \citenamefont {Porikli}, \citenamefont {Plaza},
  \citenamefont {Kehtarnavaz},\ and\ \citenamefont
  {Terzopoulos}}]{minaee_image_2022}%
  \BibitemOpen
  \bibfield  {author} {\bibinfo {author} {\bibfnamefont {S.}~\bibnamefont
  {Minaee}}, \bibinfo {author} {\bibfnamefont {Y.}~\bibnamefont {Boykov}},
  \bibinfo {author} {\bibfnamefont {F.}~\bibnamefont {Porikli}}, \bibinfo
  {author} {\bibfnamefont {A.}~\bibnamefont {Plaza}}, \bibinfo {author}
  {\bibfnamefont {N.}~\bibnamefont {Kehtarnavaz}}, \ and\ \bibinfo {author}
  {\bibfnamefont {D.}~\bibnamefont {Terzopoulos}},\ }\bibfield  {title}
  {\enquote {\bibinfo {title} {Image {Segmentation} {Using} {Deep} {Learning}:
  {A} {Survey}},}\ }\href {\doibase 10.1109/TPAMI.2021.3059968} {\bibfield
  {journal} {\bibinfo  {journal} {IEEE Transactions on Pattern Analysis and
  Machine Intelligence}\ }\textbf {\bibinfo {volume} {44}},\ \bibinfo {pages}
  {3523--3542} (\bibinfo {year} {2022})},\ \bibinfo {note} {conference Name:
  IEEE Transactions on Pattern Analysis and Machine Intelligence}\BibitemShut
  {NoStop}%
\bibitem [{\citenamefont {Alahmari}\ \emph {et~al.}(2024)\citenamefont
  {Alahmari}, \citenamefont {Goldgof}, \citenamefont {Hall},\ and\
  \citenamefont {Mouton}}]{alahmari_review_2024}%
  \BibitemOpen
  \bibfield  {author} {\bibinfo {author} {\bibfnamefont {S.~S.}\ \bibnamefont
  {Alahmari}}, \bibinfo {author} {\bibfnamefont {D.}~\bibnamefont {Goldgof}},
  \bibinfo {author} {\bibfnamefont {L.~O.}\ \bibnamefont {Hall}}, \ and\
  \bibinfo {author} {\bibfnamefont {P.~R.}\ \bibnamefont {Mouton}},\ }\bibfield
   {title} {\enquote {\bibinfo {title} {A {Review} of {Nuclei} {Detection} and
  {Segmentation} on {Microscopy} {Images} {Using} {Deep} {Learning} {With}
  {Applications} to {Unbiased} {Stereology} {Counting}},}\ }\href {\doibase
  10.1109/TNNLS.2022.3213407} {\bibfield  {journal} {\bibinfo  {journal} {IEEE
  Transactions on Neural Networks and Learning Systems}\ }\textbf {\bibinfo
  {volume} {35}},\ \bibinfo {pages} {7458--7477} (\bibinfo {year} {2024})},\
  \bibinfo {note} {conference Name: IEEE Transactions on Neural Networks and
  Learning Systems}\BibitemShut {NoStop}%
\bibitem [{\citenamefont {Wang}\ \emph {et~al.}(2024)\citenamefont {Wang},
  \citenamefont {Zhao}, \citenamefont {Xu}, \citenamefont {Han}, \citenamefont
  {Tao}, \citenamefont {Zhou}, \citenamefont {Geng}, \citenamefont {Liu},\ and\
  \citenamefont {Ji}}]{wang_systematic_2024}%
  \BibitemOpen
  \bibfield  {author} {\bibinfo {author} {\bibfnamefont {Y.}~\bibnamefont
  {Wang}}, \bibinfo {author} {\bibfnamefont {J.}~\bibnamefont {Zhao}}, \bibinfo
  {author} {\bibfnamefont {H.}~\bibnamefont {Xu}}, \bibinfo {author}
  {\bibfnamefont {C.}~\bibnamefont {Han}}, \bibinfo {author} {\bibfnamefont
  {Z.}~\bibnamefont {Tao}}, \bibinfo {author} {\bibfnamefont {D.}~\bibnamefont
  {Zhou}}, \bibinfo {author} {\bibfnamefont {T.}~\bibnamefont {Geng}}, \bibinfo
  {author} {\bibfnamefont {D.}~\bibnamefont {Liu}}, \ and\ \bibinfo {author}
  {\bibfnamefont {Z.}~\bibnamefont {Ji}},\ }\bibfield  {title}
  {{\selectlanguage{english}\enquote {\bibinfo {title} {A systematic evaluation of
  computational methods for cell segmentation},}\ }}\href {\doibase
  10.1093/bib/bbae407} {\bibfield  {journal} {\bibinfo  {journal} {Briefings in
  Bioinformatics}\ }\textbf {\bibinfo {volume} {25}},\ \bibinfo {pages}
  {bbae407} (\bibinfo {year} {2024})}\BibitemShut {NoStop}%
\bibitem [{\citenamefont {Stringer}, \citenamefont {Michaelos},\ and\
  \citenamefont {Pachitariu}(2020)}]{stringer_cellpose_2020}%
  \BibitemOpen
  \bibfield  {author} {\bibinfo {author} {\bibfnamefont {C.}~\bibnamefont
  {Stringer}}, \bibinfo {author} {\bibfnamefont {M.}~\bibnamefont {Michaelos}},
  \ and\ \bibinfo {author} {\bibfnamefont {M.}~\bibnamefont {Pachitariu}},\
  }\href {\doibase 10.1101/2020.02.02.931238} {{\selectlanguage{english}\enquote
  {\bibinfo {title} {Cellpose: a generalist algorithm for cellular
  segmentation},}\ }} (\bibinfo {year} {2020}),\ \bibinfo {note} {pages:
  2020.02.02.931238 Section: New Results}\BibitemShut {NoStop}%
\bibitem [{\citenamefont {Pachitariu}\ and\ \citenamefont
  {Stringer}(2022)}]{pachitariu_cellpose_2022}%
  \BibitemOpen
  \bibfield  {author} {\bibinfo {author} {\bibfnamefont {M.}~\bibnamefont
  {Pachitariu}}\ and\ \bibinfo {author} {\bibfnamefont {C.}~\bibnamefont
  {Stringer}},\ }\bibfield  {title} {{\selectlanguage{english}\enquote {\bibinfo
  {title} {Cellpose 2.0: how to train your own model},}\ }}\href {\doibase
  10.1038/s41592-022-01663-4} {\bibfield  {journal} {\bibinfo  {journal}
  {Nature Methods}\ }\textbf {\bibinfo {volume} {19}},\ \bibinfo {pages}
  {1634--1641} (\bibinfo {year} {2022})},\ \bibinfo {note} {publisher: Nature
  Publishing Group}\BibitemShut {NoStop}%
\bibitem [{\citenamefont {Stringer}\ and\ \citenamefont
  {Pachitariu}(2024)}]{stringer_cellpose3_2024}%
  \BibitemOpen
  \bibfield  {author} {\bibinfo {author} {\bibfnamefont {C.}~\bibnamefont
  {Stringer}}\ and\ \bibinfo {author} {\bibfnamefont {M.}~\bibnamefont
  {Pachitariu}},\ }\bibfield  {title} {{\selectlanguage{english}\enquote {\bibinfo
  {title} {Cellpose3: one-click image restoration for improved cellular
  segmentation},}\ }}\href {\doibase 10.1101/2024.02.10.579780} {\  (\bibinfo
  {year} {2024}),\ 10.1101/2024.02.10.579780}\BibitemShut {NoStop}%
\bibitem [{\citenamefont {Pachitariu}, \citenamefont {Rariden},\ and\
  \citenamefont {Stringer}(2025)}]{cellpose_sam_2024}%
  \BibitemOpen
  \bibfield  {author} {\bibinfo {author} {\bibfnamefont {M.}~\bibnamefont
  {Pachitariu}}, \bibinfo {author} {\bibfnamefont {M.}~\bibnamefont {Rariden}},
  \ and\ \bibinfo {author} {\bibfnamefont {C.}~\bibnamefont {Stringer}},\
  }\bibfield  {title} {\enquote {\bibinfo {title} {Cellpose-sam: superhuman
  generalization for cellular segmentation},}\ }\href {\doibase
  10.1101/2025.04.28.651001} {\bibfield  {journal} {\bibinfo  {journal}
  {bioRxiv}\ } (\bibinfo {year} {2025}),\ 10.1101/2025.04.28.651001},\ \Eprint
  {http://arxiv.org/abs/https://www.biorxiv.org/content/early/2025/05/01/2025.04.28.651001.full.pdf}
  {https://www.biorxiv.org/content/early/2025/05/01/2025.04.28.651001.full.pdf}
  \BibitemShut {NoStop}%
\bibitem [{\citenamefont {Kirillov}\ \emph {et~al.}(2023)\citenamefont
  {Kirillov}, \citenamefont {Mintun}, \citenamefont {Ravi}, \citenamefont
  {Mao}, \citenamefont {Rolland}, \citenamefont {Gustafson}, \citenamefont
  {Xiao}, \citenamefont {Whitehead}, \citenamefont {Berg}, \citenamefont {Lo}
  \emph {et~al.}}]{kirillov2023segment}%
  \BibitemOpen
  \bibfield  {author} {\bibinfo {author} {\bibfnamefont {A.}~\bibnamefont
  {Kirillov}}, \bibinfo {author} {\bibfnamefont {E.}~\bibnamefont {Mintun}},
  \bibinfo {author} {\bibfnamefont {N.}~\bibnamefont {Ravi}}, \bibinfo {author}
  {\bibfnamefont {H.}~\bibnamefont {Mao}}, \bibinfo {author} {\bibfnamefont
  {C.}~\bibnamefont {Rolland}}, \bibinfo {author} {\bibfnamefont
  {L.}~\bibnamefont {Gustafson}}, \bibinfo {author} {\bibfnamefont
  {T.}~\bibnamefont {Xiao}}, \bibinfo {author} {\bibfnamefont {S.}~\bibnamefont
  {Whitehead}}, \bibinfo {author} {\bibfnamefont {A.~C.}\ \bibnamefont {Berg}},
  \bibinfo {author} {\bibfnamefont {W.-Y.}\ \bibnamefont {Lo}},  \emph
  {et~al.},\ }\bibfield  {title} {\enquote {\bibinfo {title} {Segment
  anything},}\ }in\ \href@noop {} {\emph {\bibinfo {booktitle} {Proceedings of
  the IEEE/CVF international conference on computer vision}}}\ (\bibinfo {year}
  {2023})\ pp.\ \bibinfo {pages} {4015--4026}\BibitemShut {NoStop}%
\bibitem [{\citenamefont {Ouyang}\ \emph {et~al.}(2019)\citenamefont {Ouyang},
  \citenamefont {Winsnes}, \citenamefont {Hjelmare}, \citenamefont {Cesnik},
  \citenamefont {Åkesson}, \citenamefont {Xu}, \citenamefont {Sullivan},
  \citenamefont {Dai}, \citenamefont {Lan}, \citenamefont {Jinmo},
  \citenamefont {Galib}, \citenamefont {Henkel}, \citenamefont {Hwang},
  \citenamefont {Poplavskiy}, \citenamefont {Tunguz}, \citenamefont
  {Wolfinger}, \citenamefont {Gu}, \citenamefont {Li}, \citenamefont {Xie},
  \citenamefont {Buslov}, \citenamefont {Fironov}, \citenamefont {Kiselev},
  \citenamefont {Panchenko}, \citenamefont {Cao}, \citenamefont {Wei},
  \citenamefont {Wu}, \citenamefont {Zhu}, \citenamefont {Tseng}, \citenamefont
  {Gao}, \citenamefont {Ju}, \citenamefont {Yi}, \citenamefont {Zheng},
  \citenamefont {Kappel},\ and\ \citenamefont
  {Lundberg}}]{ouyang_analysis_2019}%
  \BibitemOpen
  \bibfield  {author} {\bibinfo {author} {\bibfnamefont {W.}~\bibnamefont
  {Ouyang}}, \bibinfo {author} {\bibfnamefont {C.~F.}\ \bibnamefont {Winsnes}},
  \bibinfo {author} {\bibfnamefont {M.}~\bibnamefont {Hjelmare}}, \bibinfo
  {author} {\bibfnamefont {A.~J.}\ \bibnamefont {Cesnik}}, \bibinfo {author}
  {\bibfnamefont {L.}~\bibnamefont {Åkesson}}, \bibinfo {author}
  {\bibfnamefont {H.}~\bibnamefont {Xu}}, \bibinfo {author} {\bibfnamefont
  {D.~P.}\ \bibnamefont {Sullivan}}, \bibinfo {author} {\bibfnamefont
  {S.}~\bibnamefont {Dai}}, \bibinfo {author} {\bibfnamefont {J.}~\bibnamefont
  {Lan}}, \bibinfo {author} {\bibfnamefont {P.}~\bibnamefont {Jinmo}}, \bibinfo
  {author} {\bibfnamefont {S.~M.}\ \bibnamefont {Galib}}, \bibinfo {author}
  {\bibfnamefont {C.}~\bibnamefont {Henkel}}, \bibinfo {author} {\bibfnamefont
  {K.}~\bibnamefont {Hwang}}, \bibinfo {author} {\bibfnamefont
  {D.}~\bibnamefont {Poplavskiy}}, \bibinfo {author} {\bibfnamefont
  {B.}~\bibnamefont {Tunguz}}, \bibinfo {author} {\bibfnamefont {R.~D.}\
  \bibnamefont {Wolfinger}}, \bibinfo {author} {\bibfnamefont {Y.}~\bibnamefont
  {Gu}}, \bibinfo {author} {\bibfnamefont {C.}~\bibnamefont {Li}}, \bibinfo
  {author} {\bibfnamefont {J.}~\bibnamefont {Xie}}, \bibinfo {author}
  {\bibfnamefont {D.}~\bibnamefont {Buslov}}, \bibinfo {author} {\bibfnamefont
  {S.}~\bibnamefont {Fironov}}, \bibinfo {author} {\bibfnamefont
  {A.}~\bibnamefont {Kiselev}}, \bibinfo {author} {\bibfnamefont
  {D.}~\bibnamefont {Panchenko}}, \bibinfo {author} {\bibfnamefont
  {X.}~\bibnamefont {Cao}}, \bibinfo {author} {\bibfnamefont {R.}~\bibnamefont
  {Wei}}, \bibinfo {author} {\bibfnamefont {Y.}~\bibnamefont {Wu}}, \bibinfo
  {author} {\bibfnamefont {X.}~\bibnamefont {Zhu}}, \bibinfo {author}
  {\bibfnamefont {K.-L.}\ \bibnamefont {Tseng}}, \bibinfo {author}
  {\bibfnamefont {Z.}~\bibnamefont {Gao}}, \bibinfo {author} {\bibfnamefont
  {C.}~\bibnamefont {Ju}}, \bibinfo {author} {\bibfnamefont {X.}~\bibnamefont
  {Yi}}, \bibinfo {author} {\bibfnamefont {H.}~\bibnamefont {Zheng}}, \bibinfo
  {author} {\bibfnamefont {C.}~\bibnamefont {Kappel}}, \ and\ \bibinfo {author}
  {\bibfnamefont {E.}~\bibnamefont {Lundberg}},\ }\bibfield  {title}
  {{\selectlanguage{english}\enquote {\bibinfo {title} {Analysis of the {Human}
  {Protein} {Atlas} {Image} {Classification} competition},}\ }}\href {\doibase
  10.1038/s41592-019-0658-6} {\bibfield  {journal} {\bibinfo  {journal} {Nature
  Methods}\ }\textbf {\bibinfo {volume} {16}},\ \bibinfo {pages} {1254--1261}
  (\bibinfo {year} {2019})},\ \bibinfo {note} {publisher: Nature Publishing
  Group}\BibitemShut {NoStop}%
\bibitem [{\citenamefont {Le}\ \emph {et~al.}(2022)\citenamefont {Le},
  \citenamefont {Winsnes}, \citenamefont {Axelsson}, \citenamefont {Xu},
  \citenamefont {Mohanakrishnan~Kaimal}, \citenamefont {Mahdessian},
  \citenamefont {Dai}, \citenamefont {Makarov}, \citenamefont {Ostankovich},
  \citenamefont {Xu}, \citenamefont {Benhamou}, \citenamefont {Henkel},
  \citenamefont {Solovyev}, \citenamefont {Banić}, \citenamefont {Bošnjak},
  \citenamefont {Bošnjak}, \citenamefont {Miličević}, \citenamefont
  {Ouyang},\ and\ \citenamefont {Lundberg}}]{le_analysis_2022}%
  \BibitemOpen
  \bibfield  {author} {\bibinfo {author} {\bibfnamefont {T.}~\bibnamefont
  {Le}}, \bibinfo {author} {\bibfnamefont {C.~F.}\ \bibnamefont {Winsnes}},
  \bibinfo {author} {\bibfnamefont {U.}~\bibnamefont {Axelsson}}, \bibinfo
  {author} {\bibfnamefont {H.}~\bibnamefont {Xu}}, \bibinfo {author}
  {\bibfnamefont {J.}~\bibnamefont {Mohanakrishnan~Kaimal}}, \bibinfo {author}
  {\bibfnamefont {D.}~\bibnamefont {Mahdessian}}, \bibinfo {author}
  {\bibfnamefont {S.}~\bibnamefont {Dai}}, \bibinfo {author} {\bibfnamefont
  {I.~S.}\ \bibnamefont {Makarov}}, \bibinfo {author} {\bibfnamefont
  {V.}~\bibnamefont {Ostankovich}}, \bibinfo {author} {\bibfnamefont
  {Y.}~\bibnamefont {Xu}}, \bibinfo {author} {\bibfnamefont {E.}~\bibnamefont
  {Benhamou}}, \bibinfo {author} {\bibfnamefont {C.}~\bibnamefont {Henkel}},
  \bibinfo {author} {\bibfnamefont {R.~A.}\ \bibnamefont {Solovyev}}, \bibinfo
  {author} {\bibfnamefont {N.}~\bibnamefont {Banić}}, \bibinfo {author}
  {\bibfnamefont {V.}~\bibnamefont {Bošnjak}}, \bibinfo {author}
  {\bibfnamefont {A.}~\bibnamefont {Bošnjak}}, \bibinfo {author}
  {\bibfnamefont {A.}~\bibnamefont {Miličević}}, \bibinfo {author}
  {\bibfnamefont {W.}~\bibnamefont {Ouyang}}, \ and\ \bibinfo {author}
  {\bibfnamefont {E.}~\bibnamefont {Lundberg}},\ }\bibfield  {title}
  {{\selectlanguage{english}\enquote {\bibinfo {title} {Analysis of the {Human}
  {Protein} {Atlas} {Weakly} {Supervised} {Single}-{Cell} {Classification}
  competition},}\ }}\href {\doibase 10.1038/s41592-022-01606-z} {\bibfield
  {journal} {\bibinfo  {journal} {Nature Methods}\ }\textbf {\bibinfo {volume}
  {19}},\ \bibinfo {pages} {1221--1229} (\bibinfo {year} {2022})},\ \bibinfo
  {note} {publisher: Nature Publishing Group}\BibitemShut {NoStop}%
\bibitem [{\citenamefont {Caron}\ \emph {et~al.}(2021)\citenamefont {Caron},
  \citenamefont {Touvron}, \citenamefont {Misra}, \citenamefont {Jégou},
  \citenamefont {Mairal}, \citenamefont {Bojanowski},\ and\ \citenamefont
  {Joulin}}]{caron_emerging_2021}%
  \BibitemOpen
  \bibfield  {author} {\bibinfo {author} {\bibfnamefont {M.}~\bibnamefont
  {Caron}}, \bibinfo {author} {\bibfnamefont {H.}~\bibnamefont {Touvron}},
  \bibinfo {author} {\bibfnamefont {I.}~\bibnamefont {Misra}}, \bibinfo
  {author} {\bibfnamefont {H.}~\bibnamefont {Jégou}}, \bibinfo {author}
  {\bibfnamefont {J.}~\bibnamefont {Mairal}}, \bibinfo {author} {\bibfnamefont
  {P.}~\bibnamefont {Bojanowski}}, \ and\ \bibinfo {author} {\bibfnamefont
  {A.}~\bibnamefont {Joulin}},\ }\href {\doibase 10.48550/arXiv.2104.14294}
  {\enquote {\bibinfo {title} {Emerging {Properties} in {Self}-{Supervised}
  {Vision} {Transformers}},}\ } (\bibinfo {year} {2021}),\ \bibinfo {note}
  {arXiv:2104.14294 [cs]}\BibitemShut {NoStop}%
\bibitem [{\citenamefont {Dosovitskiy}\ \emph {et~al.}(2021)\citenamefont
  {Dosovitskiy}, \citenamefont {Beyer}, \citenamefont {Kolesnikov},
  \citenamefont {Weissenborn}, \citenamefont {Zhai}, \citenamefont
  {Unterthiner}, \citenamefont {Dehghani}, \citenamefont {Minderer},
  \citenamefont {Heigold}, \citenamefont {Gelly}, \citenamefont {Uszkoreit},\
  and\ \citenamefont {Houlsby}}]{dosovitskiy_image_2021}%
  \BibitemOpen
  \bibfield  {author} {\bibinfo {author} {\bibfnamefont {A.}~\bibnamefont
  {Dosovitskiy}}, \bibinfo {author} {\bibfnamefont {L.}~\bibnamefont {Beyer}},
  \bibinfo {author} {\bibfnamefont {A.}~\bibnamefont {Kolesnikov}}, \bibinfo
  {author} {\bibfnamefont {D.}~\bibnamefont {Weissenborn}}, \bibinfo {author}
  {\bibfnamefont {X.}~\bibnamefont {Zhai}}, \bibinfo {author} {\bibfnamefont
  {T.}~\bibnamefont {Unterthiner}}, \bibinfo {author} {\bibfnamefont
  {M.}~\bibnamefont {Dehghani}}, \bibinfo {author} {\bibfnamefont
  {M.}~\bibnamefont {Minderer}}, \bibinfo {author} {\bibfnamefont
  {G.}~\bibnamefont {Heigold}}, \bibinfo {author} {\bibfnamefont
  {S.}~\bibnamefont {Gelly}}, \bibinfo {author} {\bibfnamefont
  {J.}~\bibnamefont {Uszkoreit}}, \ and\ \bibinfo {author} {\bibfnamefont
  {N.}~\bibnamefont {Houlsby}},\ }\href {\doibase 10.48550/arXiv.2010.11929}
  {\enquote {\bibinfo {title} {An {Image} is {Worth} 16x16 {Words}:
  {Transformers} for {Image} {Recognition} at {Scale}},}\ } (\bibinfo {year}
  {2021}),\ \bibinfo {note} {arXiv:2010.11929 [cs]}\BibitemShut {NoStop}%
\bibitem [{\citenamefont {Yao}\ \emph {et~al.}(2024)\citenamefont {Yao},
  \citenamefont {Hanslovsky}, \citenamefont {Huetter}, \citenamefont
  {Hoeckendorf},\ and\ \citenamefont {Richmond}}]{yao_weakly_2024}%
  \BibitemOpen
  \bibfield  {author} {\bibinfo {author} {\bibfnamefont {H.}~\bibnamefont
  {Yao}}, \bibinfo {author} {\bibfnamefont {P.}~\bibnamefont {Hanslovsky}},
  \bibinfo {author} {\bibfnamefont {J.-C.}\ \bibnamefont {Huetter}}, \bibinfo
  {author} {\bibfnamefont {B.}~\bibnamefont {Hoeckendorf}}, \ and\ \bibinfo
  {author} {\bibfnamefont {D.}~\bibnamefont {Richmond}},\ }\bibfield  {title}
  {{\selectlanguage{english}\enquote {\bibinfo {title} {Weakly {Supervised}
  {Set}-{Consistency} {Learning} {Improves} {Morphological} {Profiling} of
  {Single}-{Cell} {Images}},}\ }}\href@noop {} {\  (\bibinfo {year}
  {2024})}\BibitemShut {NoStop}%
\bibitem [{\citenamefont {Russakovsky}\ \emph {et~al.}(2015)\citenamefont
  {Russakovsky}, \citenamefont {Deng}, \citenamefont {Su}, \citenamefont
  {Krause}, \citenamefont {Satheesh}, \citenamefont {Ma}, \citenamefont
  {Huang}, \citenamefont {Karpathy}, \citenamefont {Khosla}, \citenamefont
  {Bernstein}, \citenamefont {Berg},\ and\ \citenamefont
  {Fei-Fei}}]{russakovsky2015imagenetlargescalevisual}%
  \BibitemOpen
  \bibfield  {author} {\bibinfo {author} {\bibfnamefont {O.}~\bibnamefont
  {Russakovsky}}, \bibinfo {author} {\bibfnamefont {J.}~\bibnamefont {Deng}},
  \bibinfo {author} {\bibfnamefont {H.}~\bibnamefont {Su}}, \bibinfo {author}
  {\bibfnamefont {J.}~\bibnamefont {Krause}}, \bibinfo {author} {\bibfnamefont
  {S.}~\bibnamefont {Satheesh}}, \bibinfo {author} {\bibfnamefont
  {S.}~\bibnamefont {Ma}}, \bibinfo {author} {\bibfnamefont {Z.}~\bibnamefont
  {Huang}}, \bibinfo {author} {\bibfnamefont {A.}~\bibnamefont {Karpathy}},
  \bibinfo {author} {\bibfnamefont {A.}~\bibnamefont {Khosla}}, \bibinfo
  {author} {\bibfnamefont {M.}~\bibnamefont {Bernstein}}, \bibinfo {author}
  {\bibfnamefont {A.~C.}\ \bibnamefont {Berg}}, \ and\ \bibinfo {author}
  {\bibfnamefont {L.}~\bibnamefont {Fei-Fei}},\ }\href
  {https://arxiv.org/abs/1409.0575} {\enquote {\bibinfo {title} {Imagenet large
  scale visual recognition challenge},}\ } (\bibinfo {year} {2015}),\ \Eprint
  {http://arxiv.org/abs/1409.0575} {arXiv:1409.0575 [cs.CV]} \BibitemShut
  {NoStop}%
\bibitem [{\citenamefont {Pawlowski}\ \emph {et~al.}(2016)\citenamefont
  {Pawlowski}, \citenamefont {Caicedo}, \citenamefont {Singh}, \citenamefont
  {Carpenter},\ and\ \citenamefont {Storkey}}]{Pawlowski085118}%
  \BibitemOpen
  \bibfield  {author} {\bibinfo {author} {\bibfnamefont {N.}~\bibnamefont
  {Pawlowski}}, \bibinfo {author} {\bibfnamefont {J.~C.}\ \bibnamefont
  {Caicedo}}, \bibinfo {author} {\bibfnamefont {S.}~\bibnamefont {Singh}},
  \bibinfo {author} {\bibfnamefont {A.~E.}\ \bibnamefont {Carpenter}}, \ and\
  \bibinfo {author} {\bibfnamefont {A.}~\bibnamefont {Storkey}},\ }\bibfield
  {title} {\enquote {\bibinfo {title} {Automating morphological profiling with
  generic deep convolutional networks},}\ }\href {\doibase 10.1101/085118}
  {\bibfield  {journal} {\bibinfo  {journal} {bioRxiv}\ } (\bibinfo {year}
  {2016}),\ 10.1101/085118},\ \Eprint
  {http://arxiv.org/abs/https://www.biorxiv.org/content/early/2016/11/02/085118.full.pdf}
  {https://www.biorxiv.org/content/early/2016/11/02/085118.full.pdf}
  \BibitemShut {NoStop}%
\bibitem [{\citenamefont {Chen}\ \emph {et~al.}(2024)\citenamefont {Chen},
  \citenamefont {Pham}, \citenamefont {Wang}, \citenamefont {Doron},
  \citenamefont {Moshkov}, \citenamefont {Plummer},\ and\ \citenamefont
  {Caicedo}}]{chen2024chammibenchmarkchanneladaptivemodels}%
  \BibitemOpen
  \bibfield  {author} {\bibinfo {author} {\bibfnamefont {Z.}~\bibnamefont
  {Chen}}, \bibinfo {author} {\bibfnamefont {C.}~\bibnamefont {Pham}}, \bibinfo
  {author} {\bibfnamefont {S.}~\bibnamefont {Wang}}, \bibinfo {author}
  {\bibfnamefont {M.}~\bibnamefont {Doron}}, \bibinfo {author} {\bibfnamefont
  {N.}~\bibnamefont {Moshkov}}, \bibinfo {author} {\bibfnamefont {B.~A.}\
  \bibnamefont {Plummer}}, \ and\ \bibinfo {author} {\bibfnamefont {J.~C.}\
  \bibnamefont {Caicedo}},\ }\href {https://arxiv.org/abs/2310.19224} {\enquote
  {\bibinfo {title} {Chammi: A benchmark for channel-adaptive models in
  microscopy imaging},}\ } (\bibinfo {year} {2024}),\ \Eprint
  {http://arxiv.org/abs/2310.19224} {arXiv:2310.19224 [cs.CV]} \BibitemShut
  {NoStop}%
\bibitem [{\citenamefont {Jurafsky}\ and\ \citenamefont
  {Martin}(2025)}]{jurafsky2025speech}%
  \BibitemOpen
  \bibfield  {author} {\bibinfo {author} {\bibfnamefont {D.}~\bibnamefont
  {Jurafsky}}\ and\ \bibinfo {author} {\bibfnamefont {J.~H.}\ \bibnamefont
  {Martin}},\ }\href@noop {} {\emph {\bibinfo {title} {Speech and Language
  Processing: An Introduction to Natural Language Processing, Computational
  Linguistics, and Speech Recognition}}},\ \bibinfo {edition} {3rd}\ ed.\
  (\bibinfo  {publisher} {Pearson},\ \bibinfo {year} {2025})\ \bibinfo {note}
  {online draft available at
  \url{https://web.stanford.edu/~jurafsky/slp3/}}\BibitemShut {NoStop}%
\end{thebibliography}

\end{document}